\documentclass{article}

\PassOptionsToPackage{numbers, compress}{natbib}
 \usepackage[preprint]{neurips_2026}

\usepackage[utf8]{inputenc} 
\usepackage[T1]{fontenc}    
\usepackage{hyperref}       
\usepackage{url}            
\usepackage{booktabs}       
\usepackage{amsfonts}       
\usepackage{nicefrac}       
\usepackage{microtype}      
\usepackage{xcolor}         
\usepackage{comment}
\usepackage{amsmath}
\usepackage{algorithm}
\usepackage{algorithmic}
\usepackage{graphicx}
\usepackage{wrapfig}
\usepackage{subcaption}
\usepackage{enumitem}

\title{Beyond Partner Diversity: \\
An Influence-Based Team Steering Framework for \\
Zero-Shot Human-Machine Teaming}

\author{%
  Wei Sheng \\
  Department of Computer Science\\
  Purdue University\\
  \texttt{shengw@purdue.edu} \\
  \And
  Rohan Paleja \\
  Department of Computer Science \\
  Purdue University\\
  \texttt{rpaleja@purdue.edu} \\
}

\begin{document}

\maketitle

\begin{abstract}
While AI agents are rapidly advancing from isolated tools to interactive collaborators, data-driven human-machine teaming (HMT) methods remain costly in their reliance on human interaction data across domains, teammates, and team sizes. Zero-shot coordination (ZSC) addresses this bottleneck by simulating diverse partner populations to approximate how unseen partners might behave. However, partner coverage alone is insufficient as team settings scale and communication becomes degraded. To remedy this deficiency, we propose Influence-Based Team Steering (IBTS), a framework that uses influence shaping to incentivize agents to discover diverse, high-performing team interaction patterns and further steers ongoing trajectories toward stronger learned coordination modes. We assess IBTS on Overcooked-AI in both two-agent and three-agent settings, allowing us to test whether learned coordination structure transfers beyond dyadic interaction. Our evaluation includes simulated partners, synthetic partner-style variation, and, to our knowledge, the first 30-subject Overcooked-AI HMT study involving two real human teammates and one machine teammate. Across these evaluations, IBTS improves team performance against competing baselines, highlighting the need for scaled ZSC to combine sparse-reward coordination mechanisms with partner-variation coverage rather than relying on diversity alone.
\end{abstract}

\section{Introduction}
\label{sec:intro}

Recent surging investment in embodied machines for human-proximate work, such as Apptronik's Apollo \cite{kolodny2026apptronik}, a humanoid robot designed to work alongside people and assist them with physically demanding tasks, underscores the growing need for human-machine teaming (HMT) \cite{tomasello2009we,paleja2024designs,kontogiorgos2020towards}. 
Gathering representative human interaction data for each task offers a direct way to address many dyadic HMT challenges, but this approach becomes burdensome when even one additional human joins the team. The added teammate expands the interaction space beyond individual human--machine adaptation, requiring the machine to account for individual preferences \cite{wang2025personalization}, emerging human--human coordination, and trust-mediated task-allocation dynamics under limited communication and ambiguous intent cues \cite{obi2025trust,basappa2025mind}.
This barrier motivates our focus on zero-shot coordination (ZSC) \cite{stone2010ad} and our extension from the common one-human--one-machine setup to a two-human--one-machine setup as a minimal step toward more realistic human-group collaboration.

One standard solution to the out-of-distribution (OOD) dilemma in ZSC is to train agents against populations of simulated partners and learn best responses to them \cite{strouse2021collaborating}, with recent methods further improving robustness by amplifying partner diversity within the training population \cite{sarkar2023diverse,zhao2023maximum}. 
Yet as teammate populations broaden, partner diversity alone remains inadequate because human behavior simulation cannot be covered exhaustively \cite{charakorn2024diversity}, and best-response learning may converge toward robust but static generalist policies that are poor at sustaining the effective coordination patterns \cite{ni2026theory}.
This inspires us to seek criteria beyond diversity for identifying high-performing coordination patterns, and to strengthen the learning signal used during best-response training so that ongoing interactions can be steered toward these patterns.

\begin{figure*}[t]
    \centering
    \includegraphics[width=0.87\textwidth]{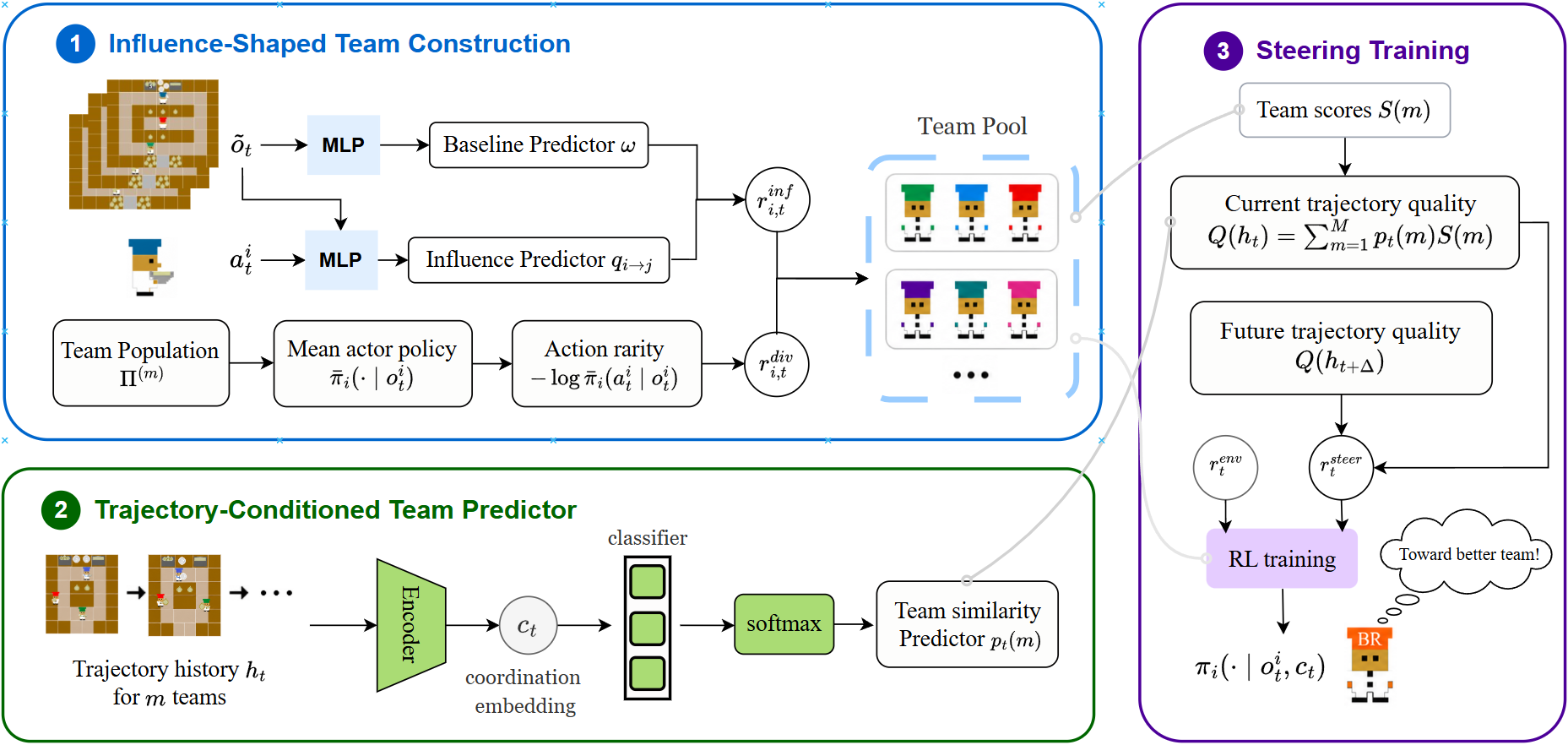}
    \caption{IBTS overview. Stage 1 constructs a diverse team pool using influence-shaped coordination rewards and behavioral diversity. Stage 2 trains a predictor that maps recent trajectory to a coordination embedding and a team-similarity distribution. Stage 3 uses the predicted team similarity and team scores to define a steering reward for training a best-response policy.}
    \label{fig:intro_ibts}
    \vspace{-5mm}
\end{figure*}
Here, we propose Influence-Based Team Steering (IBTS), a novel HMT framework that augments partner diversity with learned coordination guidance. IBTS first promotes supportive behavior patterns during team generation through influence shaping, then learns a trajectory predictor that recognizes these patterns from interaction histories. Finally, IBTS uses this real-time recognition as a steering signal, guiding the machine teammate toward stronger learned coordination patterns.

We instantiate IBTS in both standard two-agent Overcooked-AI settings~\cite{carroll2019utility} and extended three-agent Overcooked-AI settings. We evaluate IBTS with simulated learned partners, synthetic LLM partners, and real human teammates. Across these evaluations, IBTS outperforms strong diversity-focused baselines, demonstrating that learned team-performance structure provides a useful optimization signal beyond partner coverage in both 2-agent and 3-agent teams.

In summary, the contributions of this paper are:
\begin{itemize}[leftmargin=*,noitemsep]
    \item We introduce IBTS, an HMT framework that couples partner diversity with learned team-performance structure. During training, IBTS uses influence shaping to discover supportive coordination patterns and predictor-guided steering to train best-response policies. At deployment, the policy uses the learned trajectory representation to recognize the current coordination mode with unseen teammates and guide interaction toward stronger learned modes.

    \item We extend Overcooked-AI evaluation with reusable three-agent layouts and a personality-conditioned synthetic-AI protocol, showing that IBTS improves over strong diversity-focused baselines across simulated task evaluation and synthetic partner-style variation.

    \item We conduct a 30-participant human study and release, to our knowledge, the first 90-trajectory dataset of 2-human--1-AI Overcooked-AI collaboration, supporting future research on scaled HMT.
\end{itemize}

\section{Preliminaries and Related Work}
\label{sec:preliminaries}
    In this section, we review the foundations needed to contextualize IBTS, covering HMT, diversity-driven ZSC, influence shaping, and the Dec-POMDP formulation used throughout the paper.

\textbf{Human-machine teaming.}
Our work studies human-machine teaming (HMT), where autonomous agents serve as interdependent collaborators that contribute to shared objectives while adapting to human behavior during interaction \cite{o2022human,siu2021evaluation}.
A canonical testbed for studying such interaction is Overcooked-AI \cite{carroll2019utility,gessler2025overcookedv}, where the team receives a shared reward for completing cooking tasks that require coordination, alignment, and role allocation.
However, prior Overcooked-AI-based HMT evaluation centers on dyadic one-human--one-AI interaction \cite{charakorn2020investigating,fontaine2021importance,wang2024beyond}, whereas real collaborative settings often involve mixed teams in which a machine teammate must work with multiple human partners \cite{obi2025trust}.
Scaling HMT beyond dyads introduces additional complexity because the machine must adapt to individual human preferences while also preserving the coordination already emerging among human teammates, all under increasingly limited communication and ambiguous intent cues.
Data-driven HMT solutions, such as learning from human demonstrations \cite{sreeramdass2025generalized,mukherjee2022survey}, can be onerous to deploy across tasks and team compositions because representative human interaction data and task-specific interfaces are often costly or unavailable.
This burden motivates scalable HMT agents that can generalize to unseen human partners without access to target human data during training.

\textbf{Zero-Shot Coordination.}
Learning how agents can collaborate with previously unseen teammates at test time without additional adaptation is the central goal of zero-shot coordination (ZSC) \cite{bauer2008human,paleja2021utility,wang2024zsc}.
Self-play (SP), where agents learn by interacting with copies, often induces brittle conventions that generalize poorly to novel partners \cite{carroll2019utility,hu2020other}.
Population-based methods \cite{jaderberg2017population,lupu2021trajectory,xu2024population} address this limitation by exposing agents to a broader set of simulated teammates, including Fictitious Co-Play (FCP) \cite{strouse2021collaborating}, which trains agents against a population of historical partners, Maximum Entropy Population-Based Training (MEP) \cite{zhao2023maximum}, which adds an entropy bonus to promote diversity among teammate policies, and GAMMA \cite{liang2024learning}, which models heterogeneous partner behavior through generative teammate representations.
However, these methods can still struggle with coverage as the number of agents scales, because agents may adopt a broader range of role assignments, timing conventions, and interaction patterns \cite{yuan2023learning}.
Recent latent-strategy approaches \cite{hong2023learning}, such as TALENTS \cite{li2025adaptively}, learn implicit structured representations of collaborative behavior to expand the scope of diversity and outperform prior population-based baselines. Yet they remain fragile when reward feedback is too weak to reveal useful coordination behavior. 
IBTS therefore builds on latent-strategy design to span the diversity, while further injecting coordination cues beyond environment reward to reinforce high-performing pattern.

\textbf{Influence shaping.}
Generating diverse teams with coordinated behavior requires role specializations, which can emerge through cooperative multi-agent reinforcement learning (MARL) \cite{wang2020roma,liu2022heterogeneous}. A common backbone is centralized training with decentralized execution (CTDE), where centralized information can stabilize joint optimization while each agent still acts from local observations at test time \cite{lowe2017multi,amato2024introduction}. PPO-based \cite{schulman2017proximal} CTDE methods such as Multi-Agent Proximal Policy Optimization (MAPPO) \cite{yu2022surprising} have shown strong performance across multi-agent benchmarks \cite{samvelyan2019starcraft,vinyals2019grandmaster,kurach2020google}, making them a natural foundation for discovering coordinated team behavior.
However, larger teams and longer task dependencies make sparse shared rewards less informative, since rare successes may not reveal which interactions produced coordination \cite{liu2021cooperative,li2023two}.
Influence-based shaping mitigates this issue by rewarding actions that affect teammates' future behavior, thereby biasing optimization toward interaction-relevant consequences rather than isolated individual progress \cite{jaques2019social,wang2019influence}.
This mechanism can expose reusable coordination structure during self-play, but because raw influence need not be task-beneficial, IBTS links the resulting interaction patterns to learned team performance so that best-response training can steer teams toward higher-return coordination modes.

\textbf{Markov Decision Process.}
To model HMT, we formalize the cooperative task as a decentralized partially observable Markov decision process (Dec-POMDP) \cite{bernstein2002complexity,oliehoek2016concise}. A Dec-POMDP consists of a finite set of agents $N=\{1,\ldots,n\}$, global states $s\in\mathcal{S}$, per-agent action spaces $\mathcal{A}^i$ with joint action $\mathbf{a}=(a^1,\ldots,a^n)\in\mathcal{A}=\prod_i \mathcal{A}^i$, and local observations $o^i\in\Omega^i$ generated from the underlying state. At each time step $t$, each agent chooses an action $a_t^i$ using only its local information, the environment evolves according to $\mathcal{T}(s_{t+1}\mid s_t,\mathbf{a}_t)$, and the team receives a shared reward $r_t=R(s_t,\mathbf{a}_t)$. The goal is to learn decentralized policies $\pi_i(a^i\mid o^i)$ that maximize the finite-horizon discounted return
$\mathbb{E}\!\left[\sum_{t=0}^{H-1}\gamma^t r_t\right]$, where $H$ denotes the episode horizon and $\gamma \in [0,1]$ is the discount factor.

\section{Method}
\label{sec:method}
Here, we formalize the three stages of IBTS used to train a deployable machine teammate, as described in Figure~\ref{fig:intro_ibts}.
\vspace{-1mm}
\subsection{Diverse Team Pool Construction}
\label{sec:phase1}
This subsection explains the two intrinsic terms used in the team-pool construction phase. Section~\ref{sec:phase1_influence_shaping} introduces the influence-based collaboration-shaping term, which encourages stronger coordination patterns by rewarding actions that create opportunities for teammate follow-up. Section~\ref{sec:phase1_behavior_div} presents the behavioral diversity term, which helps maintain distinct conventions across teams in the pool. Algorithm~\ref{alg:team_pool_construction} summarizes the full team-pool construction procedure.

\subsubsection{Influence Shaping for Collaborative Behavior}
\label{sec:phase1_influence_shaping}
Environment reward alone may not reliably induce coordinated behavior. Appendix~\ref{app:motivation_case_study} illustrates a failure mode in a three-agent layout, where a standard MAPPO baseline struggles to discover simple assembly-line behavior. To bias the team pool toward such interaction-enabling behaviors, we augment the self-play objective with an influence-shaping term to reward actions that create opportunities for teammates to follow up.

Let $\tilde{o}_t=(o_t^i)_{i\in N}$ denote the joint observation at time $t$, and $\Phi(\tilde{o}_t)$ encode a minimal set of general collaborative features extracted from $\tilde{o}_t$, including agent locations, object locations, and object possession. Then we define a salient action $a_t^*$ as the action expected to induce the largest change in the collaborative state in the Eq.~\ref{eq:salient_action}.
\begin{equation}
\label{eq:salient_action}
a_t^*
\in
\arg\max_{a\in\mathcal{A}}\
\mathbb{E}\!\left[
\left\lVert
\Phi(\tilde{o}_{t+1})-\Phi(\tilde{o}_t)
\right\rVert_2
\;\middle|\;
\tilde{o}_t,\ a_t=a
\right]
\end{equation}

Crucially, the effect of $a_t^*$ may be delayed when teammates need to reposition or continue a multi-step chain before the induced response becomes visible. Therefore, our mechanism intentionally avoids scoring only the immediate next action and instead uses a short-horizon event label as in Eq.~\ref{eq:phase1_event_label}.
\begin{equation}
\label{eq:phase1_event_label}
y_{j,t}^{(K)}
=
\mathbf{1}
\Big\{
\exists \tau \in \{1,\dots,K\}
\text{ such that }
a_{t+\tau}^{j} = a_t^*
\Big\}
\end{equation}
Here, $y_{j,t}^{(K)}=1$ indicates that agent $j$ realizes salient action $a_t^*$ within the next $K$ steps. We use a binary indicator to capture whether the follow-up occurs at all, rather than counting repeated executions of the same target action within one window, which prevents local repetitions from being treated as separate influence events when they may simply reflect transient dynamics.

We then use the event label $y_{j,t}^{(K)}$ to define the $K$-step directed influence reward $r^{\mathrm{inf}}_{i,t}$ in Eq.~\ref{eq:phase1_influence_reward}. The reward is computed over ordered pairs of distinct agents $(i,j)$, where $i$ is the source agent receiving the intrinsic reward for shaping teammate behavior, and $j$ is the target teammate whose future follow-up event may be affected by agent $i$'s action. For each pair, we compare an observation-only baseline $\omega_j(\cdot \mid \tilde{o}_t)$ with an influence-conditioned predictor $q_{i\rightarrow j}(\cdot \mid \tilde{o}_t, a_t^i)$. The baseline estimates how much the current joint observation alone predicts the event $y_{j,t}^{(K)}$, while the influence-conditioned predictor estimates how this prediction changes after conditioning on agent $i$'s action $a_t^i$. We model their difference to estimate the contribution by $a_t^i$, and take the non-negative part so that actions are rewarded only when they increase the predicted likelihood of teammate follow-up.
\begin{equation}
\label{eq:phase1_influence_reward}
r^{\mathrm{inf}}_{i,t}
:=
\frac{1}{n-1}\!\sum_{j\neq i}
\max\!\left(
q_{i\rightarrow j}\!\left(y_{j,t}^{(K)}{=}1\mid \tilde{o}_t, a_t^i\right)
-
\omega_j\!\left(y_{j,t}^{(K)}{=}1\mid \tilde{o}_t\right),
0
\right)
\end{equation}
Both predictors are trained online from rollouts collected under the current policies. During each PPO iteration, we execute the decentralized actors to collect tuples $(\tilde{o}_t,\mathbf{a}_t,\tilde{o}_{t+1})$, construct the event label $y_{j,t}^{(K)}$ from the following $K$ steps, and update $q_{i\rightarrow j}$ and $\omega_j$ with binary cross-entropy.

The directed pairwise form in Eq.~\ref{eq:phase1_influence_reward} intentionally averages over all target teammates $j \neq i$, rather than restricting influence to spatially nearby agents or local contribution. This design matters in coordination chains, for instance in a streamline-style strategy, where an early action may immediately affect the nearest teammate, but it can also change whether a farther downstream teammate later performs a salient follow-up. Importantly, this design differs from directly rewarding a handcrafted event. As shown in Appendix~\ref{app:reward_hacking}, reward-hacking baselines that directly incentivize handoffs can produce unstable shortcuts, such as repeated local exchanges that increase shaped reward without improving task progress. This collaboration-shaping component is combined with the diversity-promoting MEP term described next.

\subsubsection{Behavioral Diversity}
\label{sec:phase1_behavior_div}
While influence shaping encourages coordination-producing behavior, it does not prevent the population from collapsing to a single dominant convention. Motivated by MEP \cite{zhao2023maximum}, we thus encourage diversity across team policies by comparing action distributions of corresponding agents across teams, rather than encouraging agents within the same team to differ from one another. To this end, we add a behavior diversity term that rewards actions that are less typical under the current population.

Let $\bar{\pi}^i(\cdot \mid o_t^i)$ denote the mean action distribution of agent index $i$ across the current population of teams. We define the behavior diversity reward as Eq.~\ref{eq:behavior_div_reward}, where $\epsilon > 0$ is a small stability constant.
\begin{equation}
\label{eq:behavior_div_reward}
r_{i,t}^{\mathrm{div}}
=
-\log\!\Big(
\max\big(
\bar{\pi}^{\,i}(a_t^i \mid o_t^i), \epsilon
\big)
\Big)
\end{equation}
This reward is large when the chosen action has low probability under the current population mean. As a result, the term encourages different teams to realize distinct conventions rather than collapsing to a single shared mode.

Finally, we integrate the influence and diversity rewards into the CTDE training objective as shown in Eq.~\ref{eq:team_pool_combined_reward}. For each agent $i$, the actor-side shaped reward combines the environment reward, the influence reward from Eq.~\ref{eq:phase1_influence_reward}, and the behavioral diversity reward from Eq.~\ref{eq:behavior_div_reward}.
\begin{equation}
\label{eq:team_pool_combined_reward}
r_{i,t}
=
r_t^{\mathrm{env}}
+
\lambda_{\mathrm{inf}} r_{i,t}^{\mathrm{inf}}
+
\lambda_{\mathrm{div}} r_{i,t}^{\mathrm{div}} 
\end{equation}

\subsection{Trajectory-Conditioned Team Predictor}
\label{sec:phase2}

For the purpose of anchoring ongoing trajectories to the learned team pool during deployment, we train a transformer-based predictor on rollout histories collected from the pool of $M$ teams. 

Let $h_t$ denote a trajectory history up to time $t$, formed from features available in agents' observations and actions, including each agent's location, facing direction, held object, and executed action. The transformer encoder $g_\psi$, parameterized by weights $\psi$, maps $h_t$ to a continuous embedding $c_t = g_\psi(h_t)$ to represent the current coordination pattern. A learned classification head with parameters $\eta=\{W_\eta,b_\eta\}$ then projects $c_t$ to a distribution over the $M$ teams as shown in Eq.~\ref{eq:trajectory_predictor_softmax}.
\begin{equation}
\label{eq:trajectory_predictor_softmax}
p_t = \mathrm{softmax}(W_\eta c_t + b_\eta) \in \mathbb{R}^M,
\end{equation}
The encoder parameters $\psi$ and classifier-head parameters $\eta$ are optimized jointly with cross-entropy on the ground-truth team label $m$ that generated history $h_t$ as shown in Eq.~\ref{eq:trajectory_predictor_loss}.
\begin{equation}
\label{eq:trajectory_predictor_loss}
\mathcal{L}_{\mathrm{pred}}
=
\mathbb{E}_{(h_t,m)}
\left[
-\log p_t(m)
\right]
\end{equation}
Optimizing Eq.~\ref{eq:trajectory_predictor_loss} encourages the transformer encoder to learn embeddings $c_t$ that distinguish among the coordination styles represented in the team pool. Intuitively, $c_t$ represents the implicit coordination pattern encoded from recent team trajectories, while $p_t(m)$ estimates how closely the ongoing trajectory history resembles each team in the pool.

\subsection{Team Steering}
\label{sec:phase3_phase4}
The goal of this stage is to learn a best-response policy $\pi(\cdot \mid o_t, c_t)$, conditioned on the learned latent pattern $c_t$, against the team pool constructed in Section~\ref{sec:phase1}. Section~\ref{sec:phase3} describes the integration of the predictor distribution $p_t(m)$ into an intrinsic steering reward, which encourages ongoing trajectories to move toward higher-quality historical coordination modes. Section~\ref{sec:phase4} presents an offline distillation step that consolidates these specialized best-response policies into a single shared policy for deployment. The full procedure is summarized in Alg.~\ref{alg:teacher_steer_distill}. 


\subsubsection{Trajectory-Quality Steering}
\label{sec:phase3}
The central idea of team steering is to reward a policy when its actions move the ongoing team trajectory closer to higher-performing teams in the learned pool. Therefore, we treat the historical team pool as a reference set of coordination modes with different quality levels. For each team $m$, we evaluate its performance over fixed-length rollouts and normalize the resulting score to obtain a quality score $S(m)\in[0,1]$. Given that $p_t(m)$ measures how strongly the current trajectory history resembles historical team $m$, we define the resulting quality-weighted trajectory score as Eq.~\ref{eq:steering_quality_score}. Intuitively, $Q(h_t)$ is a soft quality estimate of the current trajectory under the reference team pool, which is large when the current history assigns high probability to stronger historical teams.
\begin{equation}
\label{eq:steering_quality_score}
Q(h_t)
=
\sum_{m=1}^{M} p_t(m)\, S(m)
\end{equation}

We then define a short-horizon steering reward by measuring whether the trajectory moves toward a higher-quality region over the next $\Delta$ steps as Eq.~\ref{eq:steering_reward}.
\begin{equation}
\label{eq:steering_reward}
r_t^{\mathrm{steer}}
=
\max\big(Q(h_{t+\Delta}) - Q(h_t),\, 0\big)
\end{equation}
This reward is positive when the future trajectory becomes more similar to higher-quality historical coordination modes. We use the non-negative form to maintain the stability of the shaping signal because the goal of steering is to reinforce improvements toward stronger historical coordination patterns, rather than to penalize every local deviation from the current score estimate. 

Finally, we use this steering term to shape the training reward as shown in Eq.~\ref{eq:steering_total_reward}, where $r_t^{\mathrm{env}}$ is the environment reward and $\alpha$ controls the strength of the steering bonus.
\begin{equation}
\label{eq:steering_total_reward}
r_t^{\mathrm{total}}
=
r_t^{\mathrm{env}} + \alpha\, r_t^{\mathrm{steer}}
\end{equation}
In practice, for each agent index $i$, the best-response policy is trained as specialized teacher $\pi_i^{\mathrm{teacher}}$ while keeping both the trajectory predictor and the sampled partner policies from the team pool fixed. Although $r_t^{\mathrm{steer}}$ is computed from the full team trajectory, only the active teacher policy $\pi_i^{\mathrm{teacher}}$ is updated during its training loop, while the remaining partner policies are frozen. This objective thus trains each teacher not only to achieve task reward, but also to gradually shift unfolding trajectories toward coordination patterns associated with stronger historical teams.

\subsubsection{Shared-Student Distillation}
\label{sec:phase4}
Although the previous steering stage produces effective teacher policies, each teacher $\pi_i^{\mathrm{teacher}}$ is trained only as one target agent while the remaining positions are occupied by frozen partners. As a result, a single teacher may cover only the ego-centric states induced by its own agent index and partner contexts, making it insufficient as a general deployable policy. To consolidate these specialized behaviors into one policy with broader coverage, we perform an offline distillation step that pools supervision from all learned teachers.

Concretely, for each agent index $i \in \{1,\dots,n\}$, we freeze the corresponding teacher policy $\pi_i^{\mathrm{teacher}}$ and roll it out as agent $i$ while filling the remaining agent positions with sampled frozen partners from the team pool. At each timestep, we export only the acting teacher's ego-centric sample, consisting of the local observation $o_t^i$, the online latent $c_t$, and the teacher action $a_t^i$. Pooling these samples across all teacher indices and partner configurations yields the offline dataset in Eq.~\ref{eq:distill_dataset}.
\begin{equation}
\label{eq:distill_dataset}
\mathcal{D}
=
\bigcup_{i=1}^{n}
\left\{
\left(o_t^i,\, c_t,\, a_t^i\right)
\right\}
\end{equation}
Here, the superscript $i$ indicates which teacher index generated the sample, but the student itself does not take the agent index as an explicit input.

We then train a single shared student policy $\pi^{\mathrm{student}}(\cdot \mid o_t, c_t)$ by behavior cloning on the pooled dataset. The objective is standard cross-entropy over teacher actions as defined in Eq.~\ref{eq:distill_loss}.
\begin{equation}
\label{eq:distill_loss}
\mathcal{L}_{\mathrm{BC}}
=
\mathbb{E}_{(o,c,a)\sim\mathcal{D}}
\left[
-\log \pi^{\mathrm{student}}(a \mid o,c)
\right]
\end{equation}
By aggregating demonstrations from teachers trained under different indices and partner configurations, the student receives supervision over a broader set of ego-centric situations and thus serves as the final shared best-response policy at deployment.

\section{Experiments}
\label{sec:experiments}
\begin{figure}[t]
    \centering
    \includegraphics[width=0.9\linewidth]{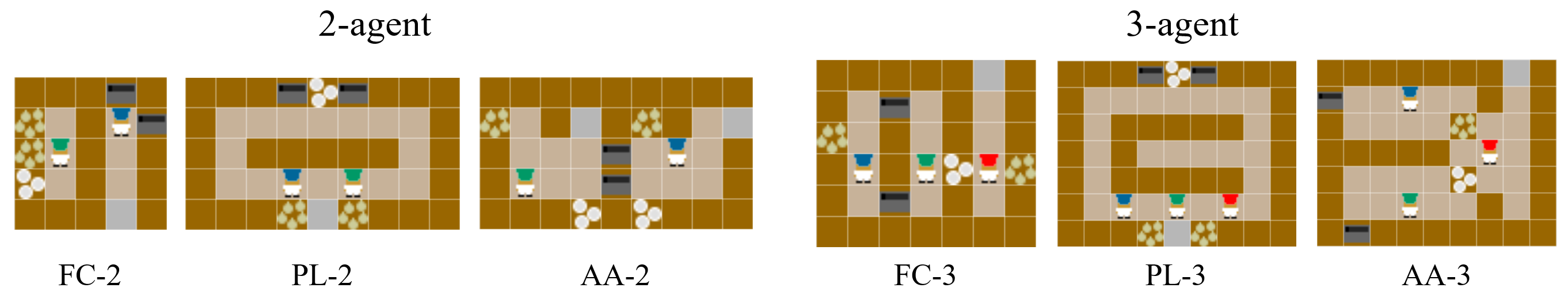}
    \caption{Overcooked-AI layouts used in the human study. We evaluate 2-agent and 3-agent settings across Forced Coordination (FC), Pipeline (PL), and Asymmetric Advantages (AA). These layouts are selected because high performance requires coordinated interaction.}
    \label{fig:hs_layout}
    \vspace{-2mm}
\end{figure}

We study whether IBTS improves coordination under broad teammate variation in Overcooked-AI layouts where high performance requires collaborative behavior rather than independent solo routines. Specifically, we use Forced Coordination (FC), Pipeline (PL), and Asymmetric Advantages (AA), as shown in Figure~\ref{fig:hs_layout}. We denote each layout by its abbreviation and team size, e.g., FC-2 refers to the 2-agent Forced Coordination setting and FC-3 refers to its 3-agent counterpart. These layouts respectively emphasize dependency resolution, sequential task flow, and uneven resource access.

In this section, we assess IBTS in simulated settings, covering both in-distribution task competence and out-of-distribution robustness to synthetic partner-style variation. Baselines include MAPPO-based self-play (SP) \cite{yu2022surprising}, FCP \cite{strouse2021collaborating}, MEP \cite{zhao2023maximum}, and GAMMA building on MEP (GAMMA) \cite{liang2024learning}, covering standard self-play and diversity-based approaches for ZSC. We discuss additional comparison-design choices in Appendix~\ref{app:comparison_design_choices}.

\subsection{In-Distribution Simulated Evaluation}
\label{subsec:in_distribution}
\begin{figure}[t]
    \centering
    \includegraphics[width=0.9\linewidth]{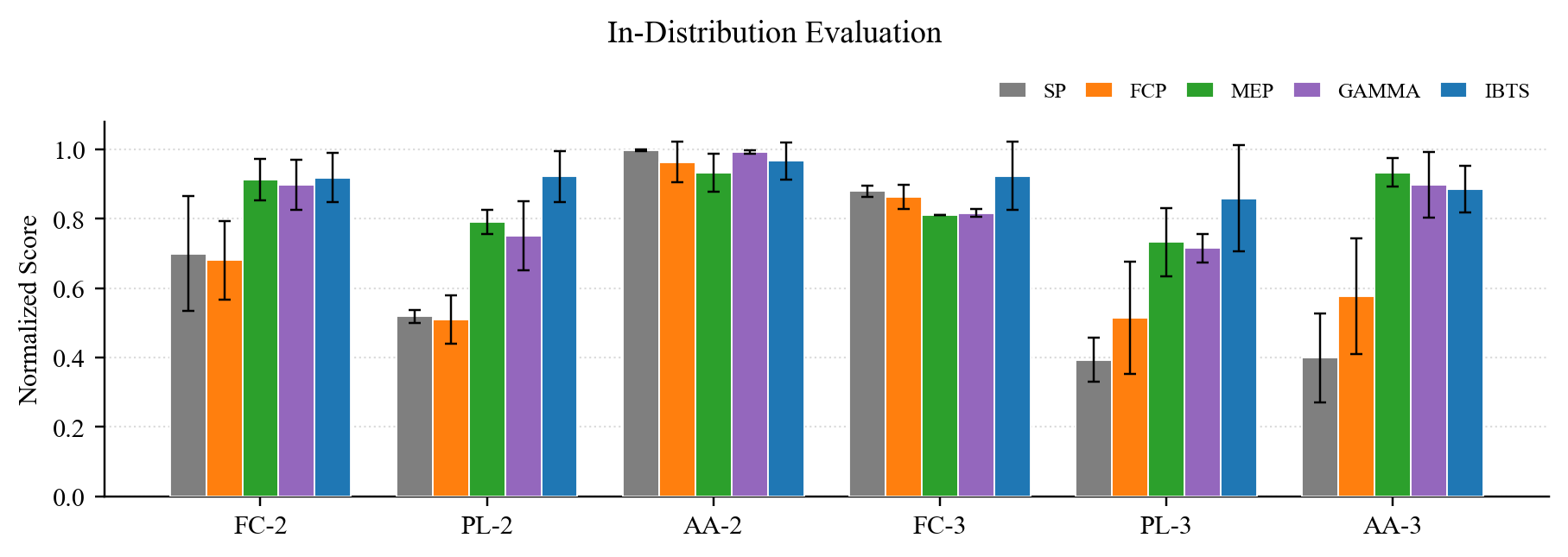}
    \caption{In-distribution simulated evaluation across 2-agent and 3-agent Overcooked layouts. Bars report mean normalized task score across 3 random seeds, with error bars showing standard deviation.}
    \label{fig:in_distribution}
    \vspace{-3mm}
\end{figure}

We first examine standard task competence under in-distribution simulation, where training and test episodes use the same layout family and agent interface without introducing new partner styles. Figure~\ref{fig:in_distribution} reports mean task scores across both 2-agent and 3-agent layouts.

The clearest gains for our framework appear in PL, where the first meaningful task reward requires agents to complete a longer coordination chain before onions reach a pot. This sparse-reward structure makes it harder for standard baselines to discover useful collaboration from task return alone. In this setting, IBTS achieves the best performance for both 2-agent and 3-agent teams, improving over MEP by $+16.5\%$ in the 2-agent case ($231.7$ vs.\ $198.8$) and by $+17.1\%$ in the 3-agent case ($148.3$ vs.\ $126.6$). By contrast, the advantage is less pronounced when early rewards are easier to discover. In AA, especially the 2-agent setting where an onion source is close to the pot, SP remains strong and achieves the highest score ($800.9$), while IBTS remains competitive ($775.7$). Overall, these results suggest that influence-based steering is most beneficial under sparse reward, where useful intermediate coordination steps may otherwise receive little direct feedback before task reward is observed, while preserving task competence when reward feedback is already more accessible.

\subsection{Synthetic LLM Partner-Style Evaluation}
\label{subsec:synthetic_llm}
\begin{figure}[t]
    \centering
    \includegraphics[width=\linewidth]{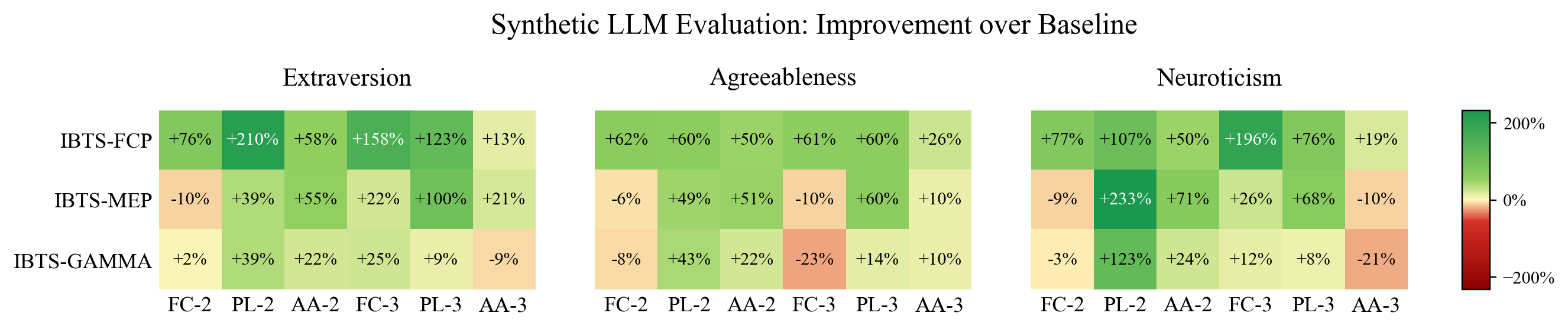}
    \caption{Synthetic LLM partner-style evaluation. Each panel corresponds to one partner personality. Columns denote layout and team-size combinations, rows denote IBTS improvements over FCP, MEP, and GAMMA, and cell values show relative improvement over the corresponding baseline. Positive values indicate improvement.}
    \label{fig:synthetic_llm}
    \vspace{-3mm}
\end{figure}

We then assess robustness to partners outside the training population. We use GPT-5-mini partners conditioned with three personality profiles: Agreeable, Extraverted, and Neurotic, following prior work showing that personality induction can produce distinct decision-making patterns in LLM-based agents~\cite{newsham2025personality}. The profiles are designed to induce different low-level coordination tendencies, such as cooperative handoffs, assertive task initiation, and cautious waiting behavior. Full game instructions, personality prompts, and the rationale for this synthetic-partner design are provided in Appendix~\ref{app:synthetic_llm_prompts}.

Figure~\ref{fig:synthetic_llm} reports relative improvement over each corresponding baseline across personality profiles, layouts, and team sizes. Although the gains are not uniform across all layout-personality pairs, IBTS improves over FCP in all 18 settings and over MEP in 13 of 18 settings. Averaged across the synthetic evaluation, IBTS achieves a mean score of $186.9$, compared with $133.7$ for MEP, $110.6$ for FCP, and $158.6$ for GAMMA. The strongest gains appear in PL, where IBTS improves over MEP by $+106.8\%$ in the 2-agent setting and $+75.8\%$ in the 3-agent setting, indicating that the method remains effective as coordination requires longer interaction chains and larger teams.

\section{Human Study}
\label{subsec:human_study}

Here, we present the details of our human-subjects study evaluating whether IBTS transfers to real human teammates. We explore the question:

\textbf{Q1.}
Does IBTS improve real-human team performance over diversity-only baselines in both dyadic HMT and two-human--one-machine group HMT?

We review the study conditions, participant protocol, evaluation measures, statistical analysis, and task-score results below.

\textbf{Study Conditions and Procedure.} We evaluate two team-size conditions: one-human--one-machine HMT and two-human--one-machine group HMT. In both conditions, participants interact with three machine partners, MEP, GAMMA, and IBTS, which were the three strongest learned-agent methods evaluated in Section~\ref{sec:experiments}. Participants played three Overcooked layouts, FC, PL, and AA, shown in Figure~\ref{fig:hs_layout}. The layout order was fixed as FC, PL, and AA, while the machine partner order was randomized.

Before the formal trials, participants received unlimited practice time in PL or AA to familiarize themselves with the controls, task mechanics, and reward structure. During the formal study, participants completed one FC--PL--AA block with a given machine partner, filled out a post-game questionnaire, and repeated this process for the remaining machine partners. Each game was capped at 400 environment timesteps and used synchronous stepping, so the environment advanced only after actions were received from all active human players and machine agents. Participants were not given information about the machine partner policies beyond the shared game rules and task objective. Verbal communication was prohibited in both team-size conditions, while natural nonverbal cues were not explicitly controlled. Full gameplay and questionnaire procedures are provided in Appendix~\ref{app:user_study_design}, Appendix~\ref{app:user_study_procedure}, and Appendix~\ref{app:user_study_gameplay}.

\textbf{Measures and Statistical Analysis.} Our primary measure is team task score, computed as the total environment reward accumulated within a 400-timestep game. For each team-size condition and layout, we compare IBTS against the learned-agent baselines using paired Wilcoxon signed-rank tests. The paired unit is the participant in the one-human--one-machine condition and the two-human team in the two-human--one-machine condition. Before conducting the nonparametric tests, we used Shapiro-Wilk tests on paired score differences to assess normality and Levene's tests with median centering to inspect variance differences.

For each layout, we test two planned comparisons: IBTS versus MEP and IBTS versus GAMMA. We apply Holm correction within each layout to account for these two comparisons. Significance markers in the result figure correspond to the Holm-corrected Wilcoxon $p$-values, with $*$ for $p<0.05$, $**$ for $p<0.01$, and $***$ for $p<0.001$. Error bars report standard error of the mean.

Participants also completed post-game questionnaires measuring perceived teammate quality. Since task performance is the primary outcome of this human study, we report the questionnaire results separately in Appendix~\ref{app:human_study_survey}.

\begin{figure*}[t]
    \centering
    \includegraphics[width=0.81\textwidth]{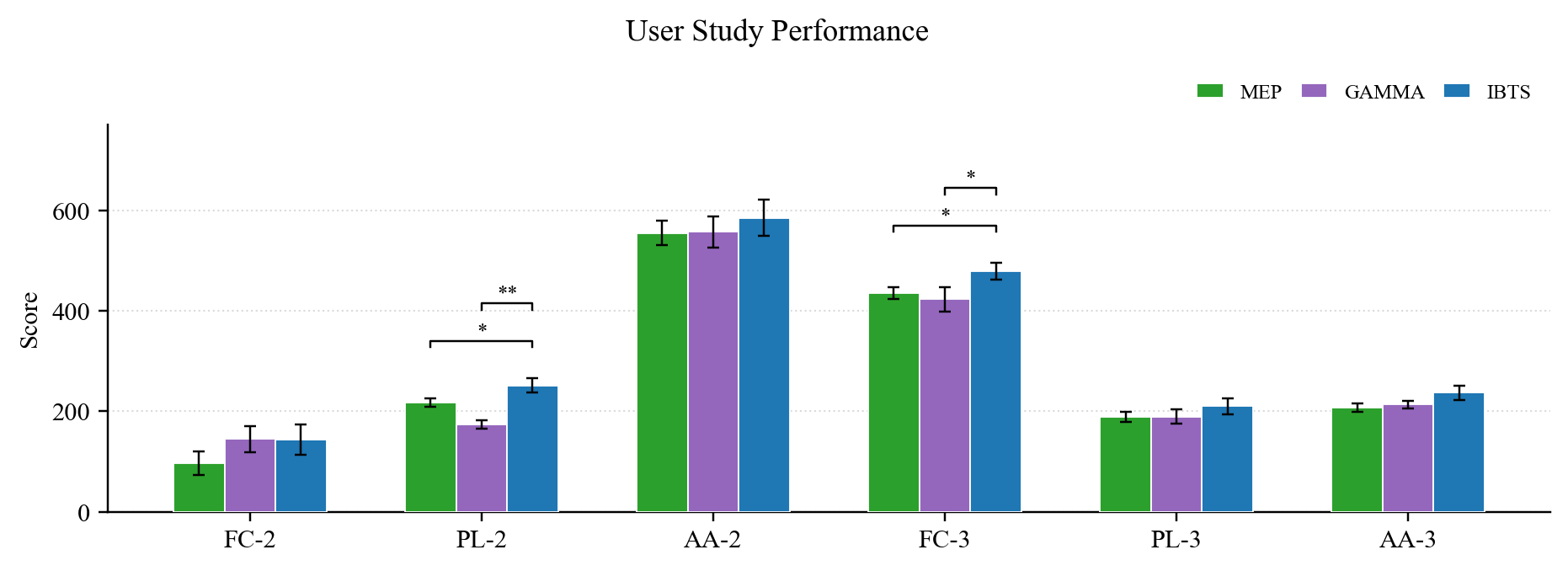}
    \caption{Human-study scores across one-human--one-machine and two-human--one-machine settings. Bars report mean team score with standard error. Significance markers denote paired Wilcoxon signed-rank tests comparing IBTS against each baseline, with Holm correction within each layout.}
    \label{fig:user_study}
    \vspace{-3mm}
\end{figure*}

\textbf{Q1 Results.}
We recruited 30 participants aged 20--35 years (mean 25.1, SD 3.3; 23 male, 7 female) under an IRB-approved protocol. Ten participants completed the one-human--one-machine condition, and the remaining 20 participants formed 10 two-human teams for the two-human--one-machine condition. We display the task-score results in Figure~\ref{fig:user_study}.

Overall, the statistically significant results support that IBTS can improve real-human team performance over learned-agent baselines in coordination-intensive settings where useful task progress is harder to discover from sparse reward alone. This pattern is consistent with the simulated and synthetic evaluations in Section~\ref{sec:experiments}, where the clearest gains also appear in settings requiring longer interaction chains. In the one-human--one-machine condition, IBTS significantly outperforms MEP in PL-2 ($251.4$ vs.\ $217.0$; Holm-corrected $p=0.039$) and significantly outperforms GAMMA in PL-2 ($251.4$ vs.\ $173.7$; Holm-corrected $p=0.0039$). In the two-human--one-machine condition, IBTS significantly outperforms both MEP and GAMMA in FC-3 ($479.1$ vs.\ $435.1$ and $423.0$, respectively; Holm-corrected $p=0.035$ for both comparisons). Together, these findings indicate that the benefits of IBTS transfer to both dyadic and group HMT.

\section{Conclusion}
\label{sec:conclusion}

This paper introduces Influence-Based Team Steering (IBTS), a framework for scalable zero-shot HMT in settings where diversity-only approaches struggle to identify useful coordination behavior under sparse rewards and longer interaction chains. IBTS mitigates this issue by using influence shaping to strengthen coordination feedback during population generation and predictor-guided steering to reuse high-performing team modes during best-response training. Across simulated, synthetic-partner, and real-human evaluations, IBTS improves over strong learned baselines in most settings, including scaled two-human--one-AI teams. 


Beyond the specific Overcooked-AI setting, our results suggest a broader design principle for scalable HMT: machines should not only be trained to tolerate diverse partners, but also to recognize and reinforce productive coordination patterns as they emerge during interaction. This perspective shifts ZSC from coverage alone toward transferable coordination structure, providing a path for using MARL-discovered behaviors to support human groups when direct human data collection is limited.

\textbf{Limitations.}
Our approach has several limitations: 1) IBTS heavily depends on the quality of the learned team-performance structure, otherwise the embedding space may not contain the cooperative behaviors that humans would prefer, limiting what the steering objective can recover. Influence shaping helps mitigate this issue by encouraging actions that induce useful teammate responses, but it does not fully solve the sparse-reward problem. For example, in PL-3, learned agents remain below the intuitive human-designed heuristic shown in Figure~\ref{fig:heuristic_gap_passing}. 2) The pairwise influence formulation in Section~\ref{sec:phase1_influence_shaping} also introduces scalability costs because it evaluates directed source-target influence terms across agent pairs. This design is intentional since IBTS aims to capture whether one agent's behavior induces useful follow-up from each teammate, including nonlocal responses that unfold downstream in a coordination chain rather than only immediate nearby interactions, but more efficient approximations will be needed for much larger teams. 3) Our human study is limited in size and scope, and the survey results in Figure~\ref{fig:survey_appendix} show that machine teammates still trail human teammates in trust, suggesting that higher task performance alone is not sufficient to close the subjective trust gap in group human-machine coordination.

\textbf{Future work.}
Future work should improve the scalability of pairwise influence estimation, extend IBTS to richer communication and continuous-control domains, and combine team steering with stronger sparse-reward MARL techniques.

\bibliographystyle{plainnat}
\bibliography{references}

@article{carroll2019utility,
  title={On the utility of learning about humans for human-AI coordination},
  author={Carroll, Micah and Shah, Rohin and Ho, Mark K and Griffiths, Tom and Seshia, Sanjit and Abbeel, Pieter and Dragan, Anca},
  journal={Advances in Neural Information Processing Systems},
  volume={32},
  year={2019}
}

@article{paleja2024designs,
  title={Designs for enabling collaboration in human-machine teaming via interactive and explainable systems},
  author={Paleja, Rohan and Munje, Michael and Chang, Kimberlee C and Jensen, Reed and Gombolay, Mathew},
  journal={Advances in Neural Information Processing Systems},
  volume={37},
  pages={64942--64969},
  year={2024}
}

@inproceedings{li2025adaptively,
  title={Adaptively coordinating with novel partners via learned latent strategies},
  author={Li, Benjamin and Shi, Shuyang and Romero, Lucia and Li, Huao and Xie, Yaqi and Kim, Woojun and Nikolaidis, Stefanos and Lewis, Charles Michael and Sycara, Katia P. and Stepputtis, Simon},
  booktitle={The Thirty-ninth Annual Conference on Neural Information Processing Systems},
  year={2025}
}

@inproceedings{kontogiorgos2020towards,
  title={Towards adaptive and least-collaborative-effort social robots},
  author={Kontogiorgos, Dimosthenis and Pelikan, Hannah R. M.},
  booktitle={Companion of the 2020 ACM/IEEE International Conference on Human-Robot Interaction},
  pages={311--313},
  year={2020}
}

@article{hong2023learning,
  title={Learning to influence human behavior with offline reinforcement learning},
  author={Hong, Joey and Levine, Sergey and Dragan, Anca},
  journal={Advances in Neural Information Processing Systems},
  volume={36},
  pages={36094--36105},
  year={2023}
}

@article{yu2022surprising,
  title={The surprising effectiveness of ppo in cooperative multi-agent games},
  author={Yu, Chao and Velu, Akash and Vinitsky, Eugene and Gao, Jiaxuan and Wang, Yu and Bayen, Alexandre and Wu, Yi},
  journal={Advances in Neural Information Processing Systems},
  volume={35},
  pages={24611--24624},
  year={2022}
}

@inproceedings{kurach2020google,
  title={Google research football: A novel reinforcement learning environment},
  author={Kurach, Karol and Raichuk, Anton and Sta{\'n}czyk, Piotr and Zaj{\k{a}}c, Micha{\l} and Bachem, Olivier and Espeholt, Lasse and Riquelme, Carlos and Vincent, Damien and Michalski, Marcin and Bousquet, Olivier and others},
  booktitle={Proceedings of the AAAI Conference on Artificial Intelligence},
  volume={34},
  pages={4501--4510},
  year={2020}
}

@article{samvelyan2019starcraft,
  title={The starcraft multi-agent challenge},
  author={Samvelyan, Mikayel and Rashid, Tabish and De Witt, Christian Schroeder and Farquhar, Gregory and Nardelli, Nantas and Rudner, Tim G. J. and Hung, Chia-Man and Torr, Philip H. S. and Foerster, Jakob and Whiteson, Shimon},
  journal={arXiv preprint arXiv:1902.04043},
  year={2019}
}

@article{strouse2021collaborating,
  title={Collaborating with humans without human data},
  author={Strouse, DJ and McKee, Kevin and Botvinick, Matt and Hughes, Edward and Everett, Richard},
  journal={Advances in Neural Information Processing Systems},
  volume={34},
  pages={14502--14515},
  year={2021}
}

@article{wang2024zsc,
  title={Zsc-eval: An evaluation toolkit and benchmark for multi-agent zero-shot coordination},
  author={Wang, Xihuai and Zhang, Shao and Zhang, Wenhao and Dong, Wentao and Chen, Jingxiao and Wen, Ying and Zhang, Weinan},
  journal={Advances in Neural Information Processing Systems},
  volume={37},
  pages={47344--47377},
  year={2024}
}

@inproceedings{hu2020other,
  title={{``Other-Play''} for zero-shot coordination},
  author={Hu, Hengyuan and Lerer, Adam and Peysakhovich, Alex and Foerster, Jakob},
  booktitle={International Conference on Machine Learning},
  pages={4399--4410},
  year={2020},
  organization={PMLR}
}

@inproceedings{zhao2023maximum,
  title={Maximum entropy population-based training for zero-shot human-ai coordination},
  author={Zhao, Rui and Song, Jinming and Yuan, Yufeng and Hu, Haifeng and Gao, Yang and Wu, Yi and Sun, Zhongqian and Yang, Wei},
  booktitle={Proceedings of the AAAI Conference on Artificial Intelligence},
  volume={37},
  pages={6145--6153},
  year={2023}
}

@article{charakorn2024diversity,
  title={Diversity is not all you need: Training a robust cooperative agent needs specialist partners},
  author={Charakorn, Rujikorn and Manoonpong, Poramate and Dilokthanakul, Nat},
  journal={Advances in Neural Information Processing Systems},
  volume={37},
  pages={56401--56423},
  year={2024}
}

@article{liang2024learning,
  title={Learning to cooperate with humans using generative agents},
  author={Liang, Yancheng and Chen, Daphne and Gupta, Abhishek and Du, Simon S and Jaques, Natasha},
  journal={Advances in Neural Information Processing Systems},
  volume={37},
  pages={60061--60087},
  year={2024}
}

@inproceedings{wang2024beyond,
  title     = {Beyond Single Stationary Policies: Meta-Task Players as Naturally Superior Collaborators},
  author    = {Wang, Haoming and Tian, Zhaoming and Song, Yunpeng and Zhang, Xiangliang and Cai, Zhongmin},
  booktitle = {Advances in Neural Information Processing Systems},
  volume    = {37},
  pages     = {78836--78862},
  year      = {2024}
}

@inproceedings{jaques2019social,
  title={Social influence as intrinsic motivation for multi-agent deep reinforcement learning},
  author={Jaques, Natasha and Lazaridou, Angeliki and Hughes, Edward and Gulcehre, Caglar and Ortega, Pedro and Strouse, DJ and Leibo, Joel Z and De Freitas, Nando},
  booktitle={International conference on machine learning},
  pages={3040--3049},
  year={2019},
  organization={PMLR}
}

@inproceedings{gessler2025overcookedv,
  title     = {OvercookedV2: Rethinking Overcooked for Zero-Shot Coordination},
  author    = {Gessler, Tobias and Dizdarevic, Tin and Calinescu, Ani and Ellis, Benjamin and Lupu, Andrei and Foerster, Jakob Nicolaus},
  booktitle = {The Thirteenth International Conference on Learning Representations},
  year      = {2025}
}

@inproceedings{fontaine2021importance,
  author    = {Fontaine, Matthew and Hsu, Ya-Chuan and Zhang, Yulun and Tjanaka, Bryon and Nikolaidis, Stefanos},
  title     = {On the Importance of Environments in Human-Robot Coordination},
  booktitle = {Proceedings of Robotics: Science and Systems},
  year      = {2021},
  address   = {Virtual},
  month     = jul,
  doi       = {10.15607/RSS.2021.XVII.038}
}

@misc{kolodny2026apptronik,
  author       = {Kolodny, Lora},
  title        = {Apptronik Raises \$520 Million to Beat Chinese Humanoids, Tesla Optimus to Market},
  year         = {2026},
  month        = feb,
  note         = {CNBC}
}

@inproceedings{wang2025personalization,
  author    = {Wang, Ruiqi and Zhao, Dezhong and Suh, Dayoon and Yuan, Ziqin and Chen, Guohua and Min, Byung-Cheol},
  title     = {Personalization in Human-Robot Interaction Through Preference-Based Action Representation Learning},
  booktitle = {Proceedings of the IEEE International Conference on Robotics and Automation},
  year      = {2025},
  pages     = {7377--7384},
  doi       = {10.1109/ICRA55743.2025.11128756}
}

@inproceedings{basappa2025mind,
  title={Mind the Gaps: How AI Shortcomings and Human Concerns May Disrupt Team Cognition in Human-AI Teams (HATs)},
  author={Basappa, Rhea and Lancaster, Caitlin and Mallick, Rohit and Flathmann, Christopher and McNeese, Nathan},
  booktitle={Proceedings of the Human Factors and Ergonomics Society Annual Meeting},
  volume={69},
  pages={354--359},
  year={2025},
  organization={SAGE Publications Sage CA: Los Angeles, CA}
}

@inproceedings{ni2026theory,
  author    = {Ni, Andrew and Stepputtis, Simon and Nikolaidis, Stefanos and Lewis, Michael and Sycara, Katia P. and Kim, Woojun},
  title     = {Theory of Mind Guided Strategy Adaptation for Zero-Shot Coordination},
  booktitle = {Proceedings of the International Conference on Autonomous Agents and Multiagent Systems},
  year      = {2026}
}

@inproceedings{stone2010ad,
  title={Ad hoc autonomous agent teams: Collaboration without pre-coordination},
  author={Stone, Peter and Kaminka, Gal and Kraus, Sarit and Rosenschein, Jeffrey},
  booktitle={Proceedings of the AAAI conference on artificial intelligence},
  volume={24},
  pages={1504--1509},
  year={2010}
}

@article{o2022human,
  title={Human--autonomy teaming: A review and analysis of the empirical literature},
  author={O’neill, Thomas and McNeese, Nathan and Barron, Amy and Schelble, Beau},
  journal={Human factors},
  volume={64},
  number={5},
  pages={904--938},
  year={2022},
  publisher={Sage Publications Sage CA: Los Angeles, CA}
}

@inproceedings{charakorn2020investigating,
  title={Investigating partner diversification methods in cooperative multi-agent deep reinforcement learning},
  author={Charakorn, Rujikorn and Manoonpong, Poramate and Dilokthanakul, Nat},
  booktitle={International Conference on Neural Information Processing},
  pages={395--402},
  year={2020},
  organization={Springer}
}

@inproceedings{obi2025trust,
  title={Investigating the Impact of Trust in Multi-Human Multi-Robot Task Allocation},
  author={Obi, Ike and Wang, Ruiqi and Jo, Wonse and Min, Byung-Cheol},
  booktitle={Proceedings of the IEEE/RSJ International Conference on Intelligent Robots and Systems (IROS)},
  address={Hangzhou, China},
  year={2025}
}

@article{mukherjee2022survey,
  title={A survey of robot learning strategies for human-robot collaboration in industrial settings},
  author={Mukherjee, Debasmita and Gupta, Kashish and Chang, Li Hsin and Najjaran, Homayoun},
  journal={Robotics and Computer-Integrated Manufacturing},
  volume={73},
  pages={102231},
  year={2022},
  publisher={Elsevier}
}

@inproceedings{lupu2021trajectory,
  title={Trajectory diversity for zero-shot coordination},
  author={Lupu, Andrei and Cui, Brandon and Hu, Hengyuan and Foerster, Jakob},
  booktitle={International Conference on Machine Learning},
  pages={7204--7213},
  year={2021},
  organization={PMLR}
}

@article{jaderberg2017population,
  title={Population based training of neural networks},
  author={Jaderberg, Max and Dalibard, Valentin and Osindero, Simon and Czarnecki, Wojciech M and Donahue, Jeff and Razavi, Ali and Vinyals, Oriol and Green, Tim and Dunning, Iain and Simonyan, Karen and others},
  journal={arXiv preprint arXiv:1711.09846},
  year={2017}
}

@article{amato2024introduction,
  title={An introduction to centralized training for decentralized execution in cooperative multi-agent reinforcement learning},
  author={Amato, Christopher},
  journal={arXiv preprint arXiv:2409.03052},
  year={2024}
}

@article{li2023two,
  title={Two heads are better than one: A simple exploration framework for efficient multi-agent reinforcement learning},
  author={Li, Jiahui and Kuang, Kun and Wang, Baoxiang and Li, Xingchen and Wu, Fei and Xiao, Jun and Chen, Long},
  journal={Advances in neural information processing systems},
  volume={36},
  pages={20038--20053},
  year={2023}
}

@inproceedings{liu2021cooperative,
  title={Cooperative exploration for multi-agent deep reinforcement learning},
  author={Liu, Iou-Jen and Jain, Unnat and Yeh, Raymond A and Schwing, Alexander},
  booktitle={International conference on machine learning},
  pages={6826--6836},
  year={2021},
  organization={PMLR}
}

@inproceedings{wang2019influence,
  author={Tonghan Wang and Jianhao Wang and Yi Wu and Chongjie Zhang},
  title={Influence-Based Multi-Agent Exploration},
  year={2020},
  booktitle={International Conference on Learning Representations}
}

@article{bernstein2002complexity,
  title={The complexity of decentralized control of Markov decision processes},
  author={Bernstein, Daniel S and Givan, Robert and Immerman, Neil and Zilberstein, Shlomo},
  journal={Mathematics of operations research},
  volume={27},
  pages={819--840},
  year={2002},
  publisher={INFORMS}
}

@book{oliehoek2016concise,
  title={A concise introduction to decentralized POMDPs},
  author={Oliehoek, Frans A and Amato, Christopher and others},
  volume={1},
  year={2016},
  publisher={Springer}
}

@inproceedings{kuba2022trust,
    title={Trust Region Policy Optimisation in Multi-Agent Reinforcement Learning},
    author={Jakub Grudzien Kuba and Ruiqing Chen and Muning Wen and Ying Wen and Fanglei Sun and Jun Wang and Yaodong Yang},
    booktitle={International Conference on Learning Representations},
    year={2022}
}

@article{schaul2015prioritized,
  title={Prioritized experience replay},
  author={Schaul, Tom and Quan, John and Antonoglou, Ioannis and Silver, David},
  journal={arXiv preprint arXiv:1511.05952},
  year={2015}
}

@article{shacklett2023extensible,
  title={An extensible, data-oriented architecture for high-performance, many-world simulation},
  author={Shacklett, Brennan and Rosenzweig, Luc Guy and Xie, Zhiqiang and Sarkar, Bidipta and Szot, Andrew and Wijmans, Erik and Koltun, Vladlen and Batra, Dhruv and Fatahalian, Kayvon},
  journal={ACM Transactions on Graphics (TOG)},
  volume={42},
  number={4},
  pages={1--13},
  year={2023},
  publisher={ACM New York, NY, USA}
}

@article{liu2022heterogeneous,
  title={Heterogeneous skill learning for multi-agent tasks},
  author={Liu, Yuntao and Li, Yuan and Xu, Xinhai and Dou, Yong and Liu, Donghong},
  journal={Advances in neural information processing systems},
  volume={35},
  pages={37011--37023},
  year={2022}
}

@inproceedings{wang2020roma,
  title     = {ROMA: Multi-Agent Reinforcement Learning with Emergent Roles},
  author    = {Wang, Tonghan and Dong, Heng and Lesser, Victor and Zhang, Chongjie},
  booktitle = {Proceedings of the 37th International Conference on Machine Learning},
  series    = {ICML},
  year      = {2020},
  publisher = {PMLR},
  pages     = {9876--9886}
}

@inproceedings{newsham2025personality,
  title     = {Personality-Driven Decision Making in LLM-Based Autonomous Agents},
  author    = {Newsham, Lewis and Prince, Daniel},
  booktitle = {Proceedings of the 24th International Conference on Autonomous Agents and Multiagent Systems},
  series    = {AAMAS},
  pages     = {1538--1547},
  year      = {2025},
  publisher = {International Foundation for Autonomous Agents and Multiagent Systems},
  address   = {Detroit, MI, USA}
}

@article{lowe2017multi,
  title={Multi-agent actor-critic for mixed cooperative-competitive environments},
  author={Lowe, Ryan and Wu, Yi I and Tamar, Aviv and Harb, Jean and Pieter Abbeel, OpenAI and Mordatch, Igor},
  journal={Advances in neural information processing systems},
  volume={30},
  year={2017}
}

@inproceedings{
    siu2021evaluation,
    title={Evaluation of Human-{AI} Teams for Learned and Rule-Based Agents in Hanabi},
    author={Ho Chit Siu and Jaime Daniel Pena and Edenna Chen and Yutai Zhou and Victor Lopez and Kyle Palko and Kimberlee Chestnut Chang and Ross Emerson Allen},
    booktitle={Advances in Neural Information Processing Systems},
    editor={A. Beygelzimer and Y. Dauphin and P. Liang and J. Wortman Vaughan},
    year={2021}
}

@article{bauer2008human,
  title={Human--robot collaboration: a survey},
  author={Bauer, Andrea and Wollherr, Dirk and Buss, Martin},
  journal={International Journal of Humanoid Robotics},
  volume={5},
  number={01},
  pages={47--66},
  year={2008},
  publisher={World Scientific}
}

@inproceedings{sreeramdass2025generalized,
  title={Generalized behavior learning from diverse demonstrations},
  author={Sreeramdass, Varshith and Paleja, Rohan R and Chen, Letian and van Waveren, Sanne and Gombolay, Matthew},
  booktitle={The Thirteenth International Conference on Learning Representations},
  year={2025}
}

@inproceedings{
sarkar2023diverse,
title={Diverse Conventions for Human-{AI} Collaboration},
author={Bidipta Sarkar and Andy Shih and Dorsa Sadigh},
booktitle={Thirty-seventh Conference on Neural Information Processing Systems},
year={2023},
}

@article{schulman2017proximal,
  title={Proximal policy optimization algorithms},
  author={Schulman, John and Wolski, Filip and Dhariwal, Prafulla and Radford, Alec and Klimov, Oleg},
  journal={arXiv preprint arXiv:1707.06347},
  year={2017}
}

@book{tomasello2009we,
  title={Why we cooperate},
  author={Tomasello, Michael},
  year={2009},
  publisher={MIT press}
}

@article{vinyals2019grandmaster,
  title={Grandmaster level in StarCraft II using multi-agent reinforcement learning},
  author={Vinyals, Oriol and Babuschkin, Igor and Czarnecki, Wojciech M and Mathieu, Micha{\"e}l and Dudzik, Andrew and Chung, Junyoung and Choi, David H and Powell, Richard and Ewalds, Timo and Georgiev, Petko and others},
  journal={nature},
  volume={575},
  number={7782},
  pages={350--354},
  year={2019},
  publisher={Nature Publishing Group UK London}
}

@inproceedings{zhang2024proagent,
  title={Proagent: building proactive cooperative agents with large language models},
  author={Zhang, Ceyao and Yang, Kaijie and Hu, Siyi and Wang, Zihao and Li, Guanghe and Sun, Yihang and Zhang, Cheng and Zhang, Zhaowei and Liu, Anji and Zhu, Song-Chun and others},
  booktitle={Proceedings of the AAAI Conference on Artificial Intelligence},
  volume={38},
  pages={17591--17599},
  year={2024}
}

@article{mccrae1999five,
  title={A five-factor theory of personality},
  author={McCrae, Robert R and Costa Jr, Paul T},
  journal={Handbook of personality: Theory and research},
  volume={2},
  number={1999},
  pages={139--153},
  year={1999},
  publisher={New York, Guilford}
}

@inproceedings{xu2024population,
  title={Population-Based Diverse Exploration for Sparse-Reward Multi-Agent Tasks.},
  author={Xu, Pei and Zhang, Junge and Huang, Kaiqi},
  booktitle={IJCAI},
  pages={283--291},
  year={2024}
}

@article{hoffman2019evaluating,
  title={Evaluating fluency in human--robot collaboration},
  author={Hoffman, Guy},
  journal={IEEE Transactions on Human-Machine Systems},
  volume={49},
  number={3},
  pages={209--218},
  year={2019},
  publisher={IEEE}
}

@article{paleja2021utility,
  title={The utility of explainable ai in ad hoc human-machine teaming},
  author={Paleja, Rohan and Ghuy, Muyleng and Ranawaka Arachchige, Nadun and Jensen, Reed and Gombolay, Matthew},
  journal={Advances in neural information processing systems},
  volume={34},
  pages={610--623},
  year={2021}
}

@inproceedings{yuan2023learning,
  title={Learning to coordinate with anyone},
  author={Yuan, Lei and Li, Lihe and Zhang, Ziqian and Chen, Feng and Zhang, Tianyi and Guan, Cong and Yu, Yang and Zhou, Zhi-Hua},
  booktitle={Proceedings of the Fifth International Conference on Distributed Artificial Intelligence},
  pages={1--9},
  year={2023}
}

\clearpage
\appendix

\section{Experimental Details and Hyperparameters}
\label{app:experimental_details}

\subsection{Reward Shaping}
\label{subsec:appendix_rewards}
In all layouts, agents must coordinate to prepare and serve onion soup under standard Overcooked dynamics.
\begin{itemize}
  \item \textbf{Sparse task reward.} Delivery of a completed three-onion soup yields $20$.
  \item \textbf{Intermediate shaping.} Placing an onion into a pot yields $+3$; picking up a cooked soup yields $+5$; dish pickup shaping is $0$.
\end{itemize}
This shaping eases exploration while preserving long-horizon coordination requirements, since high return still requires multi-step coordinated sequences to finish and deliver soups.

\subsection{Hyperparameter Rationale}
\label{subsec:appendix_hyperparam_rationale}
\begin{itemize}
  \item \textbf{Environment and rollout.}
  All training phases use $n_{\text{envs}}=64$ parallel environments and an episode horizon of $H=400$. The parallel environments provide stable rollout statistics while maintaining practical throughput, and the fixed horizon preserves delayed coordination effects. PPO collects $n_{\text{steps}}=1024$ steps per environment before each update.

  \item \textbf{PPO optimization.}
  PPO uses discount factor $\gamma=0.99$ and GAE parameter $\lambda=0.95$ to retain long-horizon returns while controlling variance. The clip range is fixed at $0.15$, the entropy coefficient is $0.05$, and the learning rate is $1\times 10^{-4}$ across the implemented IBTS training phases. We use $6$ PPO epochs per update and batch size $1024$.

  \item \textbf{Influence shaping and diversity.}
  The influence event horizon is fixed at $K=4$, providing a short future window for detecting delayed teammate follow-up. The influence coefficient is $\lambda_{\mathrm{inf}}=5.0$. The online posterior models are trained with learning rate $1\times 10^{-4}$, batch size $2048$, and one epoch per update. Behavioral diversity uses MEP coefficient $\lambda_{\mathrm{div}}=0.01$ with numerical constant $\epsilon=10^{-8}$.

  \item \textbf{Team-pool training.}
  The team pool contains $M=5$ teams. Each team is trained for $30$M environment steps using a deterministic round-robin schedule, with chunks of $5$M steps per team. The diversity objective is activated after the first full population cycle so that the population reference is based on nontrivial policies.

  \item \textbf{Trajectory predictor.}
  The team predictor uses trajectory windows of length $20$ with the features $(x,y,\text{facing},\text{held},\text{action})$ for each selected agent. It is trained as a Transformer classifier over team identifiers using cross-entropy loss. We use $d_{\mathrm{model}}=64$, $4$ attention heads, $2$ Transformer layers, feedforward dimension $128$, dropout $0.1$, batch size $256$, learning rate $10^{-3}$, weight decay $10^{-4}$, and early stopping patience $25$ over a maximum of $500$ epochs.

  \item \textbf{Predictor-guided steering.}
  Teacher training uses the frozen trajectory predictor and scalar team-quality scores from the team-pool evaluation summary. The steering reward uses coefficient $\alpha=0.5$ and temporal offset $\Delta=10$; invalid future windows receive zero steering reward. Each teacher is trained for $50$M environment steps using the same PPO settings as above.

  \item \textbf{Compute resources.}
    All models were trained on NVIDIA A30, A40, or A100 GPUs with 32GB GPU memory. Training one team pool took approximately 16 hours, and training a predictor-guided best-response agent against a fixed pool took approximately 3 days on a single GPU worker.

    \item \textbf{Implementation platform.}
    We extend the standard two-agent Overcooked-AI setting~\cite{carroll2019utility} to three-agent layouts in order to study two-human--one-machine HMT. The three-agent environments preserve the same low-level Overcooked action interface and task mechanics, including movement, interaction, object pickup and placement, pot cooking, and soup delivery. For computational efficiency, we implement these environments using the Madrona many-world simulation framework~\cite{shacklett2023extensible}. This implementation affects simulation throughput but does not change the action interface used by learned agents or human participants.
\end{itemize}

Table~\ref{tab:appendix_hparams} summarizes the primary hyperparameters used across the implemented IBTS training phases.
\begin{table}[t]
  \centering
  \caption{Key hyperparameters used across IBTS training phases.}
  \label{tab:appendix_hparams}
  \begin{small}
    \begin{sc}
      \begin{tabular}{lc}
        \toprule
        \textbf{Parameter} & \textbf{Value} \\
        \midrule
        \multicolumn{2}{c}{\textbf{Environment / rollout}} \\
        \midrule
        Parallel envs ($n_{\text{envs}}$) & 64 \\
        Episode horizon ($H$) & 400 \\
        Rollout steps ($n_{\text{steps}}$) & 1024 \\
        \midrule
        \multicolumn{2}{c}{\textbf{PPO}} \\
        \midrule
        Learning rate & $1\times 10^{-4}$ \\
        PPO epochs ($n_{\text{epochs}}$) & 6 \\
        PPO batch size & 1024 \\
        Clip range & 0.15 \\
        Discount ($\gamma$) & 0.99 \\
        GAE $\lambda$ & 0.95 \\
        Entropy coef. & 0.05 \\
        Max grad norm & 0.5 \\
        Target KL & 0.025 \\
        \midrule
        \multicolumn{2}{c}{\textbf{Influence shaping / diversity}} \\
        \midrule
        Event horizon ($K$) & 4 \\
        Influence coefficient ($\lambda_{\mathrm{inf}}$) & 5.0 \\
        Posterior learning rate & $1\times 10^{-4}$ \\
        Posterior batch size & 2048 \\
        Posterior epochs / update & 1 \\
        Posterior max grad norm & 0.5 \\

        Diversity coefficient ($\lambda_{\mathrm{div}}$) & 0.01 \\
        MEP $\epsilon$ & $1\times 10^{-8}$ \\
        \midrule
        \multicolumn{2}{c}{\textbf{Team-pool training}} \\
        \midrule
        Population size ($M$) & 5 \\
        Steps per team & $3\times 10^{7}$ \\
        Chunk steps per team & $5\times 10^{6}$ \\
        \midrule
        \multicolumn{2}{c}{\textbf{Trajectory predictor}} \\
        \midrule
        History length & 20 \\
        Input features & $x,y,\text{facing},\text{held},\text{action}$ \\
        Model type & Transformer classifier \\
        Transformer dimension ($d_{\mathrm{model}}$) & 64 \\
        Attention heads & 4 \\
        Transformer layers & 2 \\
        Feedforward dimension & 128 \\
        Dropout & 0.1 \\
        Predictor batch size & 256 \\
        Predictor epochs & 500 \\
        Predictor learning rate & $1\times 10^{-3}$ \\
        Weight decay & $1\times 10^{-4}$ \\
        Train / val / test split & $0.8 / 0.1 / 0.1$ \\
        \midrule
        \multicolumn{2}{c}{\textbf{Predictor-guided steering}} \\
        \midrule
        Steering coefficient ($\alpha$) & 0.5 \\
        Temporal offset ($\Delta$) & 10 \\
        \bottomrule
      \end{tabular}
    \end{sc}
  \end{small}
\end{table}

\subsection{Choice of Intrinsic Scaling Coefficients}
\label{sec:appendix_intrinsic_coefficients}

The influence and diversity terms are integrated as intrinsic shaping rewards in the team-pool construction objective in Eq.~\ref{eq:team_pool_combined_reward}. The coefficients $\lambda_{\mathrm{inf}}$ and $\lambda_{\mathrm{div}}$ control the relative strength of the coordination and diversity incentives. We set $\lambda_{\mathrm{inf}}=5.0$ and $\lambda_{\mathrm{div}}=0.01$.

Our choice of $\lambda_{\mathrm{inf}}=5.0$ is primarily a reward-scale calibration. The influence reward in Eq.~\ref{eq:phase1_influence_reward} is a baseline-adjusted probability lift, averaged over target teammates and computed over a short-horizon coordination event. Because it is formed as a difference between predicted event probabilities and then averaged across targets, the per-step influence term is naturally small, typically on the order of $10^{-3}$ after averaging over $j\neq i$. Multiplying by $\lambda_{\mathrm{inf}}=5.0$ keeps the shaped influence contribution on the order of a few $\times 10^{-3}$, which is large enough to provide coordination guidance when task reward is sparse, but small enough to remain secondary once environment reward becomes informative.

This scale choice is especially important early in training. At initialization, deliveries are rare and the average per-step environment reward can be very small, so a modest influence term can help reinforce salient actions that increase the probability of short-horizon teammate follow-up. As learning begins, the extrinsic reward rises into a regime where task progress becomes more frequent, while the scaled influence term remains at an auxiliary scale. Thus, $\lambda_{\mathrm{inf}}=5.0$ provides a meaningful coordination bias during the sparse-reward phase without turning influence maximization into the primary optimization target.

For the behavioral diversity term, we use $\lambda_{\mathrm{div}}=0.01$ because the unscaled reward in Eq.~\ref{eq:behavior_div_reward} typically lies between roughly $0.5$ and $1.7$ in our runs. Scaling this term by $0.01$ yields an auxiliary contribution of approximately $0.005$ to $0.017$, which is comparable to the scaled influence contribution and small relative to task reward once training becomes productive. This keeps diversity as a regularizer that discourages collapse to a single convention without overwhelming the environment objective or the influence-shaping signal.

For predictor-guided steering, we use $\alpha=0.5$. The team-quality scores $S(m)$ are normalized to lie in $[0,1]$, and the predictor distribution $p_t(m)$ is also normalized over the reference team pool. Therefore, the soft trajectory-quality estimate $Q(h_t)=\sum_{m=1}^{M} p_t(m)S(m)$ also lies in $[0,1]$. The steering reward in Eq.~\ref{eq:steering_reward} is based on the short-horizon improvement $Q(h_{t+\Delta})-Q(h_t)$, so its raw magnitude is usually much smaller than the absolute quality score and is often on the order of $10^{-2}$. Multiplying by $\alpha=0.5$ keeps the steering contribution at a comparable auxiliary scale to the influence and diversity shaping terms, while preventing the teacher from optimizing the predictor score at the expense of task reward.

\subsection{Sensitivity to Event Horizon $K$}
\label{subsec:ablations_kstep}

\begin{wrapfigure}{r}{0.45\columnwidth}
  \centering
  \includegraphics[width=0.45\columnwidth]{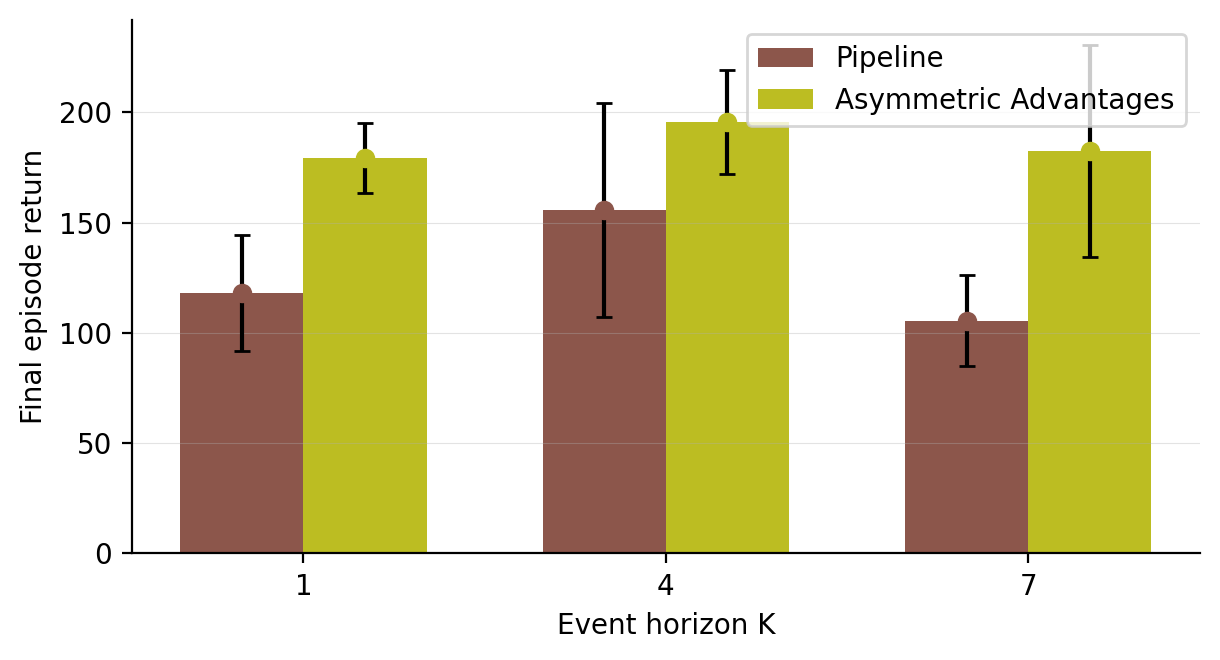}
  \caption{
    Sensitivity of MAPPO+IS to the event horizon $K$ on PL-3 and AA-3.
    Bars show mean final episode return over 12 seeds, and error bars show $\pm 1$ standard deviation.
    }
  \label{fig:kstep_sensitivity_mappo_is}
  \vspace{-3mm}
\end{wrapfigure}

We ablate the event horizon $K$ in the event-level influence label while holding the MAPPO+IS pipeline fixed. We compare $K\in\{1,4,7\}$ on the 3-agent Pipeline (PL-3) and Asymmetric Advantages (AA-3) layouts, ranging from immediate next-step labeling to a longer timing tolerance.

Figure~\ref{fig:kstep_sensitivity_mappo_is} shows that a medium horizon improves performance over immediate next-step labeling in both layouts. In Pipeline, moving from $K{=}1$ to $K{=}4$ yields a $+31.7\%$ gain in mean return, while extending to $K{=}7$ produces a $-32.1\%$ drop relative to $K{=}4$. In Asymmetric Advantages, $K{=}4$ improves mean return by $+9.1\%$ over $K{=}1$, and $K{=}7$ reduces performance by $-6.7\%$ relative to $K{=}4$.

These trends match the timing structure of Overcooked coordination. With $K{=}1$, many legitimate follow-ups cannot occur on the next step because teammates may need to move, avoid collisions, or reach an interaction tile before responding. A short future window therefore gives the influence label enough tolerance to capture delayed teammate responses. However, overly long horizons can dilute attribution: as $K$ increases, positive labels become less selective and may occur under typical dynamics even when the initiating action did not meaningfully cause the follow-up. This weakens directed influence estimation and is consistent with the drop at $K{=}7$, especially in Pipeline.

We therefore use $K{=}4$ as the default in the main experiments, balancing tolerance to short coordination delays with keeping the event label tied to the initiating action. Overall, these results suggest that influence shaping is most effective when the event horizon captures near-term teammate responses without making the label too permissive.

\subsection{Comparison Design Choices}
\label{app:comparison_design_choices}

We exclude TALENTS \cite{li2025adaptively} although it is a closer recent baseline that outperforms MEP and GAMMA, because its pre-designed high-level action space introduces task-specific abstractions, whereas our comparisons keep all methods on the same low-level Overcooked action interface. This choice avoids giving one method additional task knowledge through hand-designed abstractions.

For the same reason, during best-response training we sample frozen partners uniformly from the team pool rather than using priority-based partner sampling \cite{schaul2015prioritized}. In MEP-style best-response training, partners that are harder to collaborate with can be assigned higher sampling priority \cite{zhao2023maximum}, shifting training away from a pure average-case objective and toward a smooth approximation of maximizing worst-case performance over the partner population. While this can improve robustness, it also introduces an additional design choice about which partners should be emphasized during training. In our setting, such curated partner exposure could inject extra prior knowledge into the comparison. We therefore use uniform partner sampling to keep the evaluation focused on the predictor-guided steering objective rather than on a manually shaped partner-selection curriculum.

\section{Training Algorithms}
This section provides implementation-level pseudocode for the main IBTS training procedures. Algorithm~\ref{alg:team_pool_construction} summarizes influence-shaped team-pool construction, corresponding to Section~\ref{sec:phase1}. Algorithm~\ref{alg:teacher_steer_distill} summarizes predictor-guided teacher learning and shared-student distillation, corresponding to Sections~\ref{sec:phase3_phase4}.

\label{app:algorithms}
\begin{algorithm}[h]
  \caption{Team-Pool Construction with Influence Shaping and Behavioral Diversity}
  \label{alg:team_pool_construction}
  \begin{algorithmic}[1]
  \small
    \STATE Initialize $M$ teams $\{\Pi^{(m)}\}_{m=1}^{M}$, where each team is $\Pi^{(m)}=\{\pi_1^{(m)},\ldots,\pi_n^{(m)}\}$ with $n$ agents
    \WHILE{not converged}
      \STATE Sample a team $\Pi$, keep remaining teams frozen
      \FOR{each policy update}
        \STATE Sample joint action $\mathbf{a}_t \sim \Pi(\cdot \mid \tilde{o}_t)$ at each timestep $t$, step environment and collect rollout
        \FOR{each ordered pair $(i,j)$ with $j \neq i$}
            \STATE Construct $y_{j,t}^{(K)}$ using Eq.~\ref{eq:phase1_event_label}, update $q_{i\rightarrow j}$ and $\omega_j$ using binary cross-entropy on the rollout
        \ENDFOR
        \STATE Compute $r_{i,t}^{\mathrm{inf}}$ using Eq.~\ref{eq:phase1_influence_reward} and $r_{i,t}^{\mathrm{div}}$ using Eq.~\ref{eq:behavior_div_reward}
        \FOR{each agent $i \in \{1,\dots,n\}$ and timestep $t$}
          \STATE Form actor reward $r_{i,t}$ using Eq.~\ref{eq:team_pool_combined_reward}
        \ENDFOR
        \STATE Reduce per-agent rewards to team reward $r_t^{\mathrm{team}}=\frac{1}{n}\sum_{i=1}^{n}r_{i,t}$
        \STATE Compute GAE / returns for actor and critic, and update only the active team $\Pi$
      \ENDFOR
    \ENDWHILE
  \end{algorithmic}
\end{algorithm}

\begin{algorithm}[h]
  \caption{Predictor-Guided Teacher Learning and Shared-Student Distillation}
  \label{alg:teacher_steer_distill}
  \begin{algorithmic}[1]
  \small
    \FOR{$\pi_i^{\mathrm{teacher}}$ where $i \in \{1,\dots,n\}$}
      \WHILE{not converged}
          \STATE Compute predictor outputs $c_t$ and $p_t(m)$ from recent history $h_t$ at each timestep $t$
          \STATE Sample target action $a_t^i \sim \pi_i^{\mathrm{teacher}}(\cdot \mid o_t^i, c_t)$, sample frozen-partner actions for the remaining agent positions, step environment, and collect rollout
          \STATE Compute $Q(h_t)$ using Eq.~\ref{eq:steering_quality_score}, $r_t^{\mathrm{steer}}$ using Eq.~\ref{eq:steering_reward}, and $r_t^{\mathrm{total}}$ using Eq.~\ref{eq:steering_total_reward}
          \STATE Compute GAE / returns for actor and critic, and update only $\pi_i^{\mathrm{teacher}}$
      \ENDWHILE
    \ENDFOR
    \STATE Roll out $\pi_i^{\mathrm{teacher}}$ as agent $i$ with sampled frozen partners and export $o_t^i, c_t, a_t^i$ to $\mathcal{D}$ in Eq.~\ref{eq:distill_dataset}
    \WHILE{not converged}
      \STATE Sample minibatches from $\mathcal{D}$
      \STATE Update shared student policy $\pi^{\mathrm{student}}$ using Eq.~\ref{eq:distill_loss}
    \ENDWHILE
  \end{algorithmic}
\end{algorithm}

\section{Ablation and Case Study on Influence Shaping}
\label{app:influence_case_study}

This appendix provides additional evidence for the influence-shaping component used to construct the self-play team pool. Throughout this section, we refer to the influence-shaping variant as \textbf{IS}.

\subsection{Motivating Case Study in Three-Agent Overcooked-AI}
\label{app:motivation_case_study}

We include this case study to motivate the use of influence shaping for constructing the self-play team pool. The goal is not to argue that hand-designed heuristics are a practical solution, but rather to illustrate that standard cooperative MARL objectives can fail to discover simple interaction patterns even when such patterns are clearly beneficial. This motivates adding an explicit shaping signal that encourages agents to create opportunities for teammate follow-up behavior.

\begin{figure}[t]
  \centering
  \begin{minipage}[t]{0.3\linewidth}
    \centering
    \includegraphics[width=\linewidth]{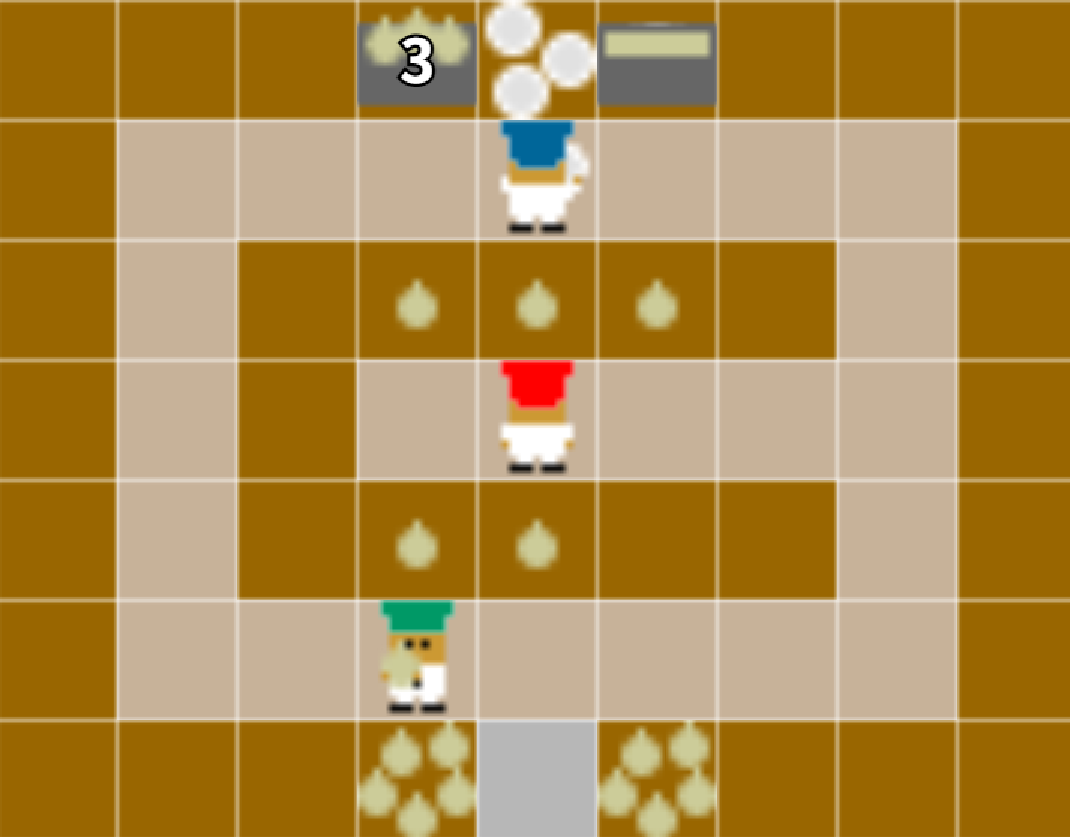}
    \caption*{\small (a) Pipeline layout with passing-style heuristic behavior}
  \end{minipage}
  \hspace{0.1\linewidth}
  \begin{minipage}[t]{0.38\linewidth}
    \centering
    \includegraphics[width=\linewidth]{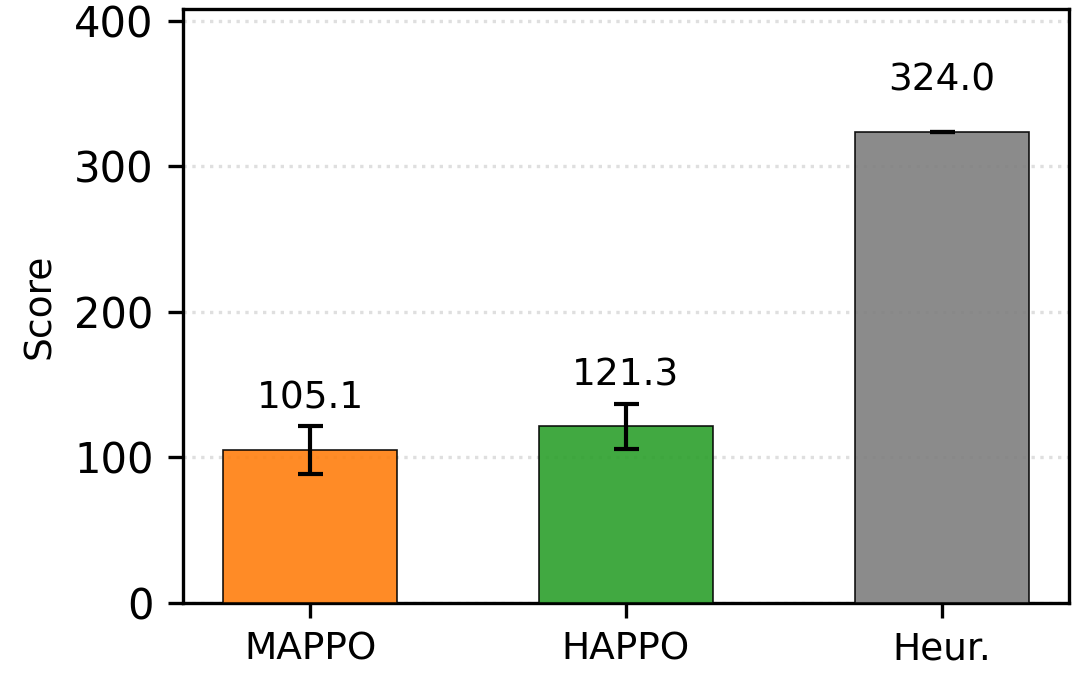}
    \caption*{\small (b) Standard MARL baselines underperform the heuristic}
  \end{minipage}

  \caption{Case study showing that standard cooperative MARL baselines can struggle to discover passing-style coordination in a three-agent Pipeline layout.}
  \label{fig:heuristic_gap_passing}
\end{figure}

Figure~\ref{fig:heuristic_gap_passing} shows a simple three-agent collaborative scenario, which we refer to as the Pipeline layout. The layout admits an intuitive passing-style strategy in which agents specialize into complementary roles and move task objects through the workspace in a coordinated sequence. Although this behavior is simple to specify as a heuristic, it requires agents to learn that an action may be valuable because it enables a teammate's later response, rather than because it immediately increases reward.

In this setting, standard CTDE baselines such as MAPPO and Heterogeneous-Agent Proximal Policy Optimization (HAPPO) \cite{kuba2022trust} fail to match the hand-designed passing heuristic, achieving 68\% and 63\% lower reward, respectively. This gap suggests that sparse shared reward alone may be insufficient to induce the interaction structure required for efficient collaboration, even in relatively simple multi-agent layouts. In particular, agents must discover not only useful individual actions, but also the temporal dependencies between one agent's behavior and another agent's subsequent follow-up.

\subsection{Scaling Across Two-, Three-, and Four-Agent Layouts}
\label{app:layout_scaling}

\begin{figure*}[t]
  \centering

  \begin{minipage}[t]{0.302\textwidth}
    \centering
    \includegraphics[width=\linewidth]{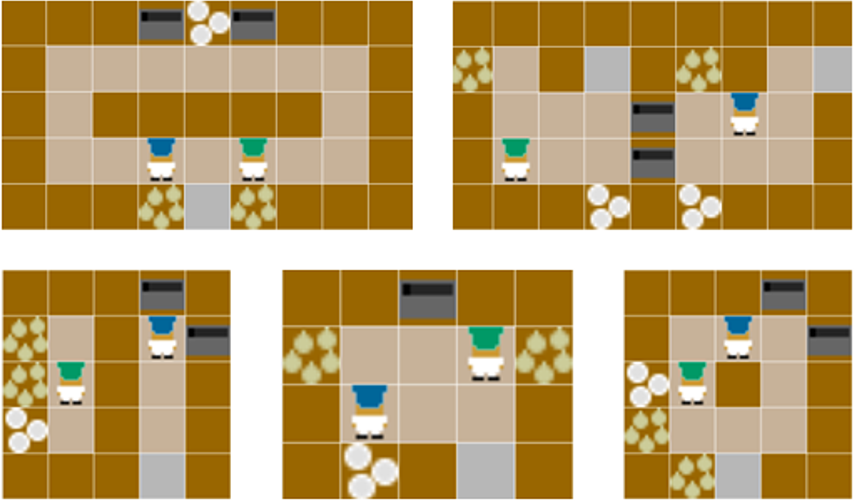}\\
    {\small 2-agent}
  \end{minipage}\hspace{7mm}
  \begin{minipage}[t]{0.345\textwidth}
    \centering
    \includegraphics[width=\linewidth]{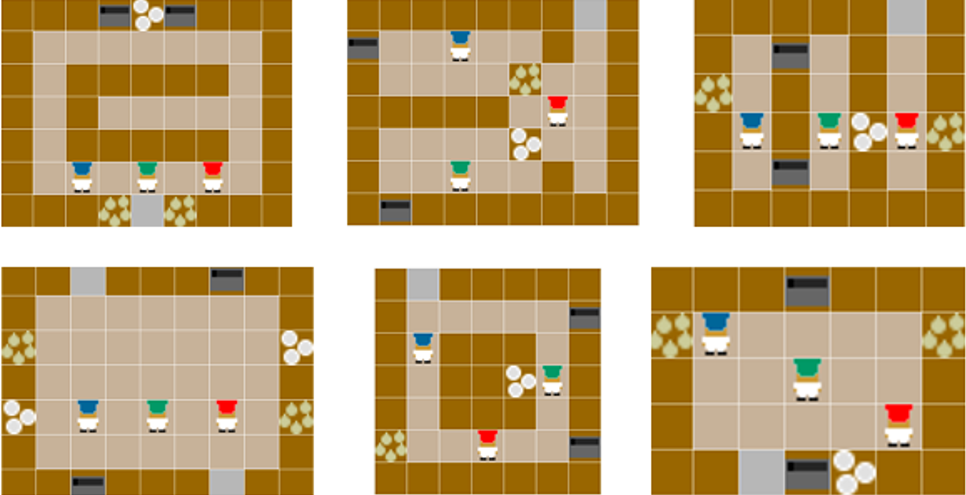}\\
    {\small 3-agent}
  \end{minipage}\hspace{7mm}
  \begin{minipage}[t]{0.206\textwidth}
    \centering
    \includegraphics[width=\linewidth]{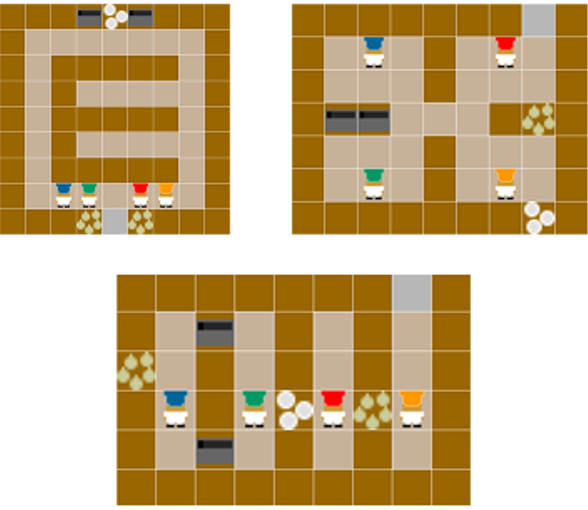}\\
    {\small 4-agent}
  \end{minipage}

  \caption{
    Layout overview for the 2-agent, 3-agent, and 4-agent settings.
    In the 2-agent setting, the top row (left to right) shows Pipeline and Asymmetric Advantages, and the bottom row (left to right) shows Forced Coordination, Cramped Room, and Coordination Ring.
    In the 3-agent setting, the top row (left to right) shows Pipeline, Asymmetric Advantages, and Forced Coordination, and the bottom row (left to right) shows Open Room, Coordination Ring, and Cramped Room.
    In the 4-agent setting, the top row (left to right) shows Pipeline, Asymmetric Advantages, and Forced Coordination.
  }
  \label{fig:layout_overview}
\end{figure*}

To examine whether the influence-shaping signal remains useful beyond a single case study, we evaluate it across two-, three-, and four-agent Overcooked-AI layouts. Figure~\ref{fig:layout_overview} summarizes the layouts used in these settings, and Table~\ref{tab:all_results_table} reports the full numerical results across all layouts and team sizes.

\begin{table}[t]
    \caption{Mean $\pm$ std over 12 seeds with maximum in parentheses for all layouts and team sizes where best mean is bolded.}
    \label{tab:all_results_table}
  \begin{center}
    \begin{small}
      \setlength{\tabcolsep}{3.2pt}      
      \renewcommand{\arraystretch}{1.05} 
      \resizebox{\textwidth}{!}{%
        \begin{tabular}{lcccccc}
          \toprule
          Layout & MAPPO+IS & HAPPO+IS & MAPPO & HAPPO & FCP & SP \\
          \midrule
          \multicolumn{7}{c}{\textbf{2-agent settings}} \\
          \midrule
          Pipeline & $\mathbf{222.7 \pm 32.1\ (290)}$ & $197.2 \pm 46.6\ (247)$ & $191.2 \pm 65.9\ (284)$ & $183.3 \pm 65.6\ (290)$ & $73.7 \pm 22.3\ (106)$ & $112.8 \pm 1.9\ (118)$ \\
          Asymmetric Advantages & $542.4 \pm 24.9\ (590)$ & $586.4 \pm 14.0\ (596)$ & $530.2 \pm 26.2\ (584)$ & $540.8 \pm 17.2\ (567)$ & $337.2 \pm 27.2\ (363)$ & $596.0 \pm 0.0\ (596)$ \\
          Forced Coordination & $304.2 \pm 47.9\ (358)$ & $\mathbf{307.7 \pm 41.9\ (349)}$ & $280.8 \pm 47.7\ (349)$ & $288.8 \pm 44.7\ (352)$ & $191.2 \pm 22.7\ (216)$ & $287.8 \pm 49.0\ (355)$ \\
          Cramped Room & $310.1 \pm 8.2\ (315)$ & $312.0 \pm 3.1\ (315)$ & $306.5 \pm 12.1\ (315)$ & $314.0 \pm 2.3\ (315)$ & $245.0 \pm 3.5\ (247)$ & $312.2 \pm 9.8\ (315)$ \\
          Coordination Ring & $266.8 \pm 15.5\ (287)$ & $268.2 \pm 12.6\ (281)$ & $265.0 \pm 13.9\ (284)$ & $274.1 \pm 7.4\ (281)$ & $183.7 \pm 13.5\ (207)$ & $276.2 \pm 11.5\ (284)$ \\
          \midrule
          \multicolumn{7}{c}{\textbf{3-agent settings}} \\
          \midrule
          Pipeline & $\mathbf{155.5 \pm 48.4\ (272)}$ & $136.6 \pm 29.7\ (207)$ & $105.1 \pm 16.8\ (134)$ & $121.3 \pm 15.3\ (150)$ & $55.5 \pm 7.8\ (67)$ & $119.1 \pm 35.6\ (204)$ \\
          Asymmetric Advantages & $195.6 \pm 23.5\ (252)$ & $\mathbf{200.8 \pm 15.5\ (218)}$ & $176.7 \pm 26.9\ (213)$ & $191.8 \pm 17.5\ (213)$ & $117.7 \pm 20.4\ (144)$ & $188.3 \pm 34.3\ (221)$ \\
          Forced Coordination & $330.7 \pm 38.3\ (391)$ & $\mathbf{337.6 \pm 44.6\ (417)}$ & $283.9 \pm 31.7\ (380)$ & $304.8 \pm 44.2\ (383)$ & $197.0 \pm 16.0\ (213)$ & $322.1 \pm 36.5\ (360)$ \\
          Open Room & $344.6 \pm 16.4\ (374)$ & $360.4 \pm 11.6\ (380)$ & $318.7 \pm 50.5\ (377)$ & $353.4 \pm 17.8\ (386)$ & $202.3 \pm 22.4\ (221)$ & $379.8 \pm 52.7\ (423)$ \\
          Coordination Ring & $213.5 \pm 10.3\ (244)$ & $209.5 \pm 7.7\ (218)$ & $206.6 \pm 10.9\ (218)$ & $203.8 \pm 18.1\ (244)$ & $166.6 \pm 15.0\ (187)$ & $262.5 \pm 9.4\ (278)$ \\
          Cramped Room & $432.0 \pm 30.9\ (485)$ & $452.9 \pm 28.4\ (493)$ & $388.1 \pm 52.3\ (451)$ & $424.8 \pm 59.9\ (491)$ & $271.8 \pm 26.9\ (345)$ & $504.3 \pm 28.6\ (553)$ \\
          \midrule
          \multicolumn{7}{c}{\textbf{4-agent settings}} \\
          \midrule
          Pipeline & $112.0 \pm 22.0\ (139)$ & $\mathbf{128.3 \pm 23.7\ (155)}$ & $97.8 \pm 20.6\ (136)$ & $120.6 \pm 21.5\ (147)$ & $66.9 \pm 16.4\ (95)$ & $88.1 \pm 25.2\ (112)$ \\
          Asymmetric Advantages & $229.8 \pm 18.8\ (247)$ & $\mathbf{253.4 \pm 26.2\ (286)}$ & $197.8 \pm 35.3\ (244)$ & $225.5 \pm 36.0\ (281)$ & $171.9 \pm 16.1\ (204)$ & $173.8 \pm 50.2\ (218)$ \\
          Forced Coordination & $314.4 \pm 24.8\ (352)$ & $\mathbf{387.5 \pm 48.6\ (423)}$ & $282.0 \pm 22.5\ (349)$ & $373.3 \pm 56.6\ (420)$ & $176.6 \pm 17.9\ (204)$ & $348.3 \pm 37.8\ (402)$ \\
          \bottomrule
        \end{tabular}%
      }
    \end{small}
  \end{center}
  \vskip -0.1in
\end{table}

\subsection{Reward-Hacking Ablation}
\label{app:reward_hacking}

A natural alternative to influence shaping is to directly reward a hand-designed coordination event. In Overcooked-AI, the most obvious such event is a handoff, where one agent drops an item and another agent picks it up within a short temporal window. To separate the contribution of influence shaping from task-specific reward engineering, we evaluate a reward-hacking baseline that adds a dense handoff bonus directly to the environment reward and trains the resulting policy with MAPPO.

\begin{table}[t]
\centering
\caption{Reward-hacking ablation in three-agent Overcooked-AI. Results show mean $\pm$ standard deviation over 12 random seeds.}
\label{tab:reward_hacking}
\begin{tabular}{lccc}
\toprule
Layout & MAPPO+IS & MAPPO & Reward-hacking \\
\midrule
Pipeline & $155.5 \pm 48.4$ & $105.1 \pm 16.8$ & $48.0 \pm 47.7$ \\
Asymmetric Advantages & $195.6 \pm 23.5$ & $176.7 \pm 26.9$ & $107.7 \pm 80.5$ \\
\bottomrule
\end{tabular}
\end{table}

Table~\ref{tab:reward_hacking} shows that directly rewarding handoffs does not reliably improve task performance. The high variance of the reward-hacking baseline suggests unstable behavior across seeds. Some runs can become trapped in poor local optima, where agents repeatedly exchange items to obtain dense immediate reward without improving task progress, while other runs largely ignore the added bonus.

This result highlights an important distinction between handcrafted event rewards and influence shaping. The reward-hacking baseline optimizes the frequency of one predefined local event, whereas IS rewards actions that increase the likelihood of task-relevant teammate follow-up behavior. As a result, IS can support a broader class of downstream coordination patterns, such as placing ingredients into pots or reorganizing teammate roles, without hard-coding a specific handoff behavior.

\subsection{Generality Beyond Overcooked-AI}
\label{app:grf_generality}
We also include a preliminary generality check in Google Research Football (GRF). Unlike Overcooked-AI, coordination in GRF cannot be reduced to a single event type. Depending on the game state, useful teammate responses may include short passes, long passes, shots, or off-ball repositioning. We use a setting with three controlled attackers against three defenders and one goalkeeper, with attackers initialized closer to midfield to make the strategy more varied.

Using the same MAPPO backbone and 30M training steps, we evaluate over 10 deterministic episodes. Passing ratio denotes the fraction of possession actions that are pass actions, and ball hold time denotes the total number of timesteps in which the controlled team holds the ball without passing.

\begin{table}[t]
\centering
\caption{GRF evaluation of influence shaping.}
\label{tab:grf_is}
\begin{tabular}{lccc}
\toprule
Method & Goals & Passing ratio & Ball hold time \\
\midrule
MAPPO & 1/10 & 0.0630 & 564 \\
MAPPO+IS & 7/10 & 0.0904 & 322 \\
\bottomrule
\end{tabular}
\end{table}

As shown in Table~\ref{tab:grf_is}, MAPPO+IS scores more goals and increases the passing ratio relative to MAPPO. It also reduces ball hold time without passing by 42.9\%, which is consistent with a policy that relies more on teammate interaction and less on prolonged single-agent dribbling. While this experiment is preliminary, it suggests that influence shaping can provide a useful coordination bias beyond Overcooked-AI.

\section{Synthetic LLM Evaluation Prompts}
\label{app:synthetic_llm_prompts}

This section describes the prompt structure used for the personality-conditioned synthetic LLM partner evaluation. Each LLM partner receives a shared game manual followed by a personality-specific behavior prompt. The model is instructed to output exactly one low-level Overcooked action from \{\texttt{north}, \texttt{south}, \texttt{east}, \texttt{west}, \texttt{stay}, \texttt{interact}\} at each timestep, with no explanation or extra text.

\textbf{Shared game manual.}
The shared game manual describes the Overcooked task objective, reward structure, action semantics, object types, and interaction rules. It specifies that agents should work with teammates to maximize team reward by preparing and serving onion soup. The reward structure is: $+3$ for placing an onion into a pot, $+5$ for taking cooked soup out of a pot using a dish, and $+20$ for serving a completed cooked three-onion soup. The manual also explains that agents act in the same kitchen, can hold only one object at a time, must face an object or station before using \texttt{interact}, and may use movement actions to change facing direction even when movement is blocked.

The manual provides task guidance but does not prescribe a complete high-level strategy. In particular, it tells the model to make useful progress toward higher reward, avoid repetitive non-progressing behavior, use dishes when pots are cooking or ready, and choose an action consistent with its assigned personality when multiple actions appear reasonable. This design keeps the LLM partner at the low-level action-selection interface rather than giving it an explicit planner or skill hierarchy.

\textbf{Agreeable partner.}
The Agreeable profile follows the Five-Factor view of agreeableness as cooperative and empathetic \cite{mccrae1999five}. The prompt describes the partner as cooperative, considerate, accommodating, generous, and willing to support others. It instructs the agent to pay close attention to teammates, prefer actions that support smooth teamwork, help through handoffs or continuation of teammate-started work, give way when another teammate is already committed to an object or route, avoid blocking or duplicating effort, and choose complementary tasks when teammates are already making useful progress.

\textbf{Extraverted partner.}
The Extraverted profile describes the partner as active, assertive, energetic, and comfortable taking initiative. The prompt instructs the agent to act early, commit quickly to useful tasks, prefer actions that directly move the task forward, keep momentum, use handoffs when clearly efficient, and take the lead in ambiguous situations rather than only reacting to teammates. When several actions appear similarly useful, the agent is encouraged to choose the more assertive, faster, and more directly productive action.

\textbf{Neurotic partner.}
The Neurotic profile describes the partner as anxious, cautious, hesitant, and sensitive to uncertainty. The prompt instructs the agent to be less likely to commit quickly when the best action is unclear, prefer safer and more certain actions, avoid uncertain handoffs or tightly timed coordination unless intent is clear, hesitate before taking initiative without an established coordination pattern, and choose actions that are less risky, less disruptive, and easier to control. The prompt still instructs the agent to attend to teammates and the team objective, but allows uncertainty to make the partner slower to commit.

\textbf{Purpose of the synthetic profiles.}
The synthetic profiles are not intended to create optimal LLM teammates. Instead, they provide controlled low-level partner variation that complements the real-human HMT evaluation. We use personality-conditioned GPT-5-mini partners to induce distinct action-selection tendencies, such as cooperative handoffs, assertive task initiation, and cautious waiting behavior, while preserving the same game manual, action space, and observation interface across partners. This follows prior work showing that personality induction can produce systematic differences in LLM-based agent decision making~\cite{newsham2025personality}. We do not use skill-conditioned LLM agents such as ProAgent~\cite{zhang2024proagent}, because our goal is not to construct an explicitly planned high-skill teammate with additional task-knowledge or planning mechanisms. Instead, the synthetic profiles are designed to simulate limited-communication HMT settings in which the machine teammate must infer partner intent from observed actions. This lets us evaluate whether IBTS remains robust when teammate behavior differs from the learned-agent population, while keeping the evaluation focused on partner-style variation rather than on the capabilities of a separately engineered LLM planner.

\section{Human Study Protocol}
\label{app:user_study_protocol}

\subsection{Study Design and Conditions}
\label{app:user_study_design}

The study used a within-condition design crossing three Overcooked layouts with three AI partners. Each participant completed nine formal games in total. The layouts were Forced Coordination (FC), Pipeline (PL), and Asymmetric Advantages (AA), with layout order fixed as FC, PL, and AA. The AI partners were MEP, GAMMA, and IBTS, and their order was randomized to reduce ordering effects. Each game was capped at 400 environment timesteps.

We evaluated two team-size conditions. The 2-agent condition consisted of one human participant and one AI teammate. The 3-agent condition consisted of two human participants and one AI teammate. Thus, the 3-agent condition evaluates two-human--one-AI collaboration under the same layout and AI-partner structure as the 2-agent condition.

\subsection{Study Procedure}
\label{app:user_study_procedure}

Before the formal trials, participants were given unlimited practice time in PL or AA to familiarize themselves with the controls, task mechanics, and reward structure. During the formal study, participants completed one FC--PL--AA block with a given AI partner, filled out a team-effectiveness questionnaire, and then repeated the same process for the remaining AI partners. Thus, each questionnaire response summarizes the participant's experience with one AI partner across all three layouts in the corresponding team-size condition.

The study was conducted under an IRB-approved protocol with 30 participants in total. Most sessions were completed in person: 28 participants completed the study in person, while one two-human team, consisting of 2 participants, completed the study online through remote control of the study laptop.

A single game typically lasted between three and six minutes depending on human decision time, and a full session lasted approximately one hour. Each participant received a \$10 Amazon gift card after completing the study.

\subsection{Gameplay Constraints}
\label{app:user_study_gameplay}

The game used synchronous stepping: the environment advanced only after actions were received from all active human players and AI agents. This prevented AI agents from moving faster than humans simply because the model could act more quickly.

To approximate degraded or limited-communication settings, verbal communication was prohibited during gameplay in both the 2-agent and 3-agent conditions. This design makes the study a test of implicit coordination, since participants must infer machine behavior, and in the 3-agent condition the other human's intent, from observed actions alone.

\subsection{Questionnaire Materials}
\label{app:user_study_questionnaires}

\begin{figure*}[t]
    \centering
    \begin{subfigure}[t]{0.48\textwidth}
        \centering
        \includegraphics[width=\linewidth]{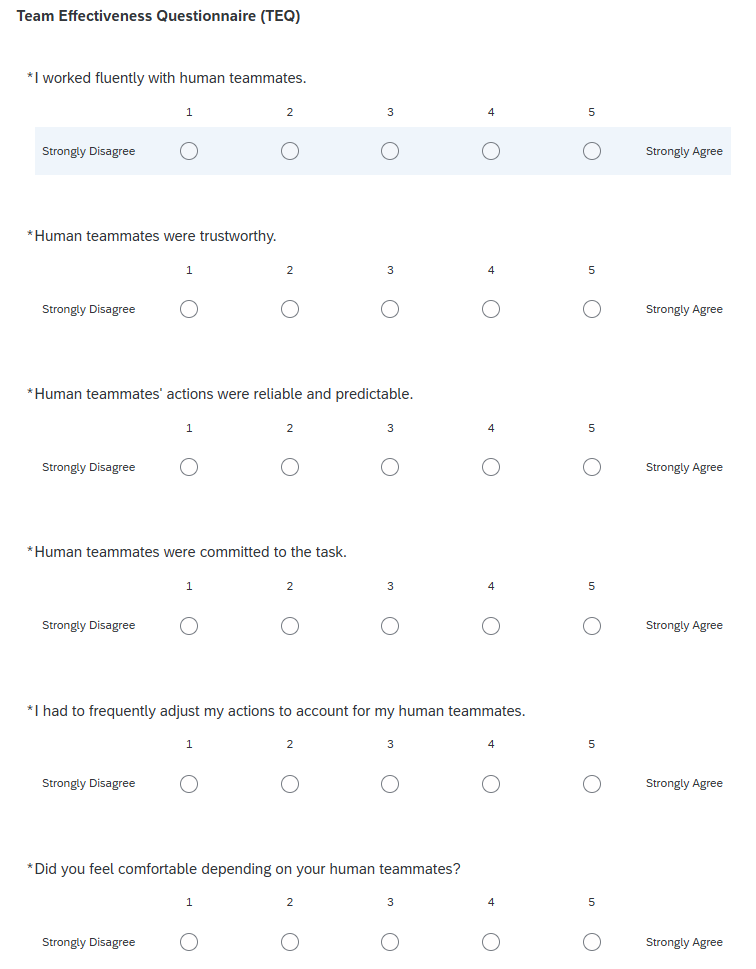}
        \caption{Page 1.}
    \end{subfigure}
    \hfill
    \begin{subfigure}[t]{0.48\textwidth}
        \centering
        \includegraphics[width=\linewidth]{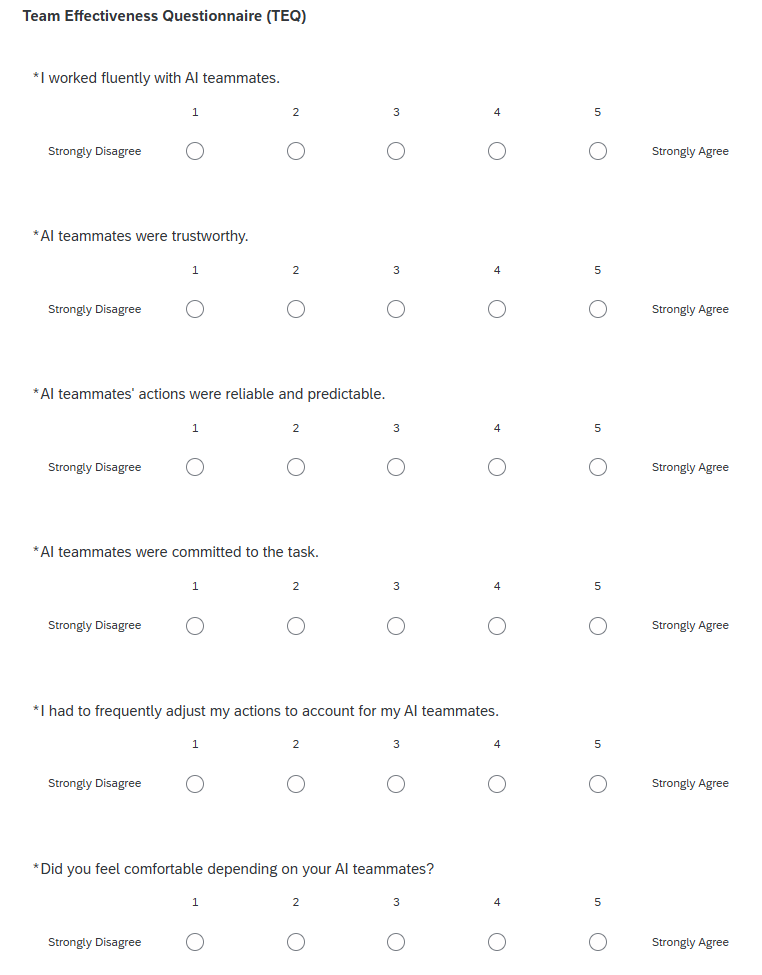}
        \caption{Page 2.}
    \end{subfigure}

    \vspace{0.8em}

    \begin{subfigure}[t]{0.48\textwidth}
        \centering
        \includegraphics[width=\linewidth]{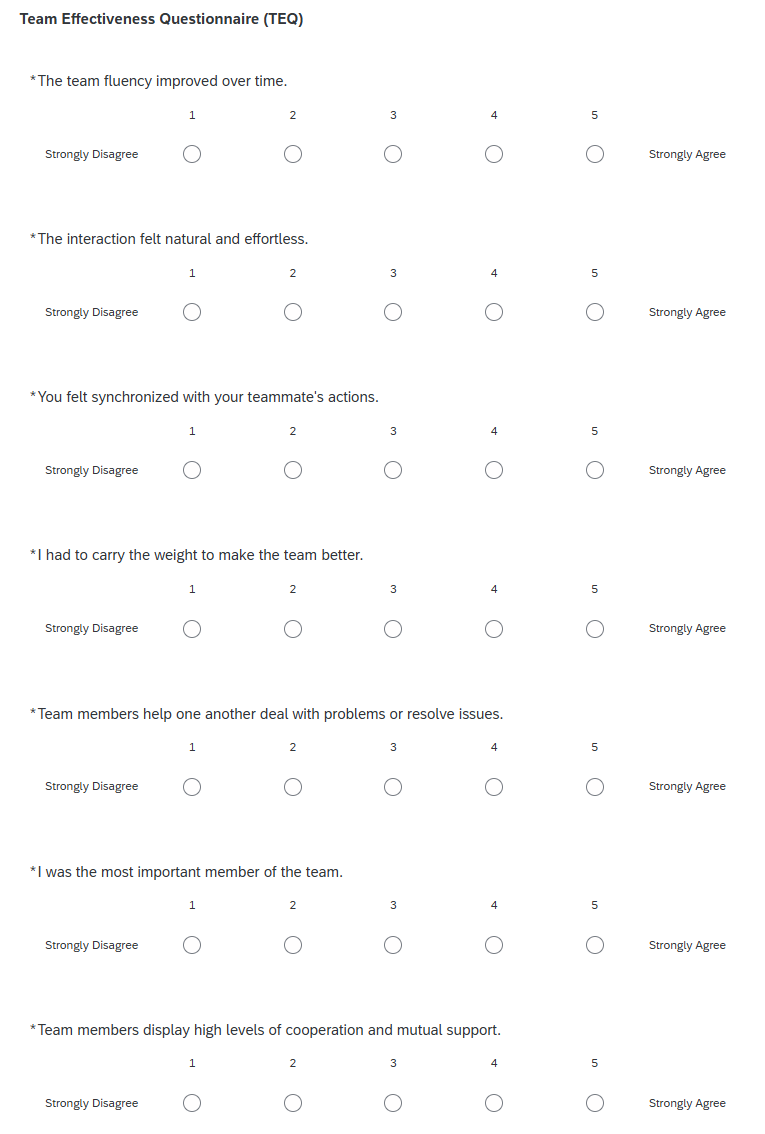}
        \caption{Page 3.}
    \end{subfigure}
    \hfill
    \begin{subfigure}[t]{0.48\textwidth}
        \centering
        \includegraphics[width=\linewidth]{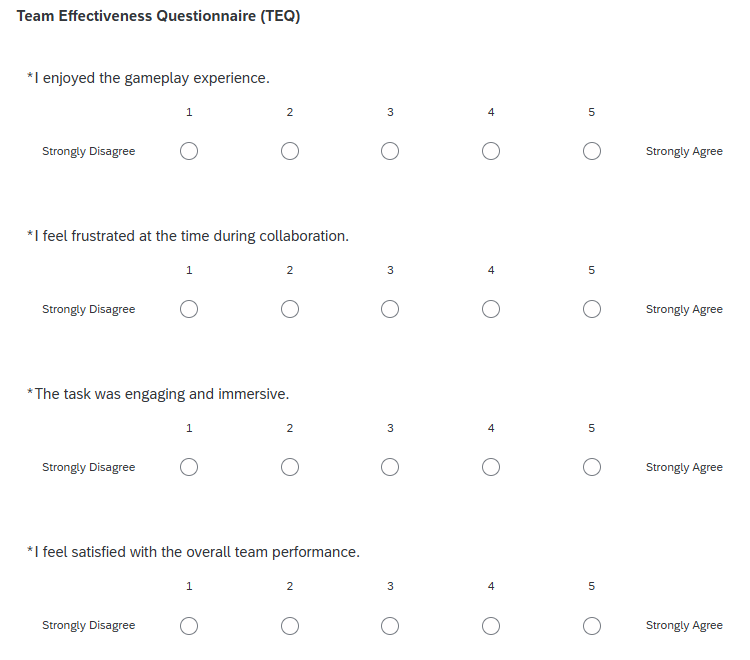}
        \caption{Page 4.}
    \end{subfigure}

    \caption{Team-effectiveness questionnaire used in the human study. Items measure perceived fluency, trust, work balance, and satisfaction on 5-point Likert scales.}
    \label{fig:appendix_teq}
\end{figure*}

\begin{figure*}[t]
    \centering
    \begin{subfigure}[t]{0.48\textwidth}
        \centering
        \includegraphics[width=\linewidth]{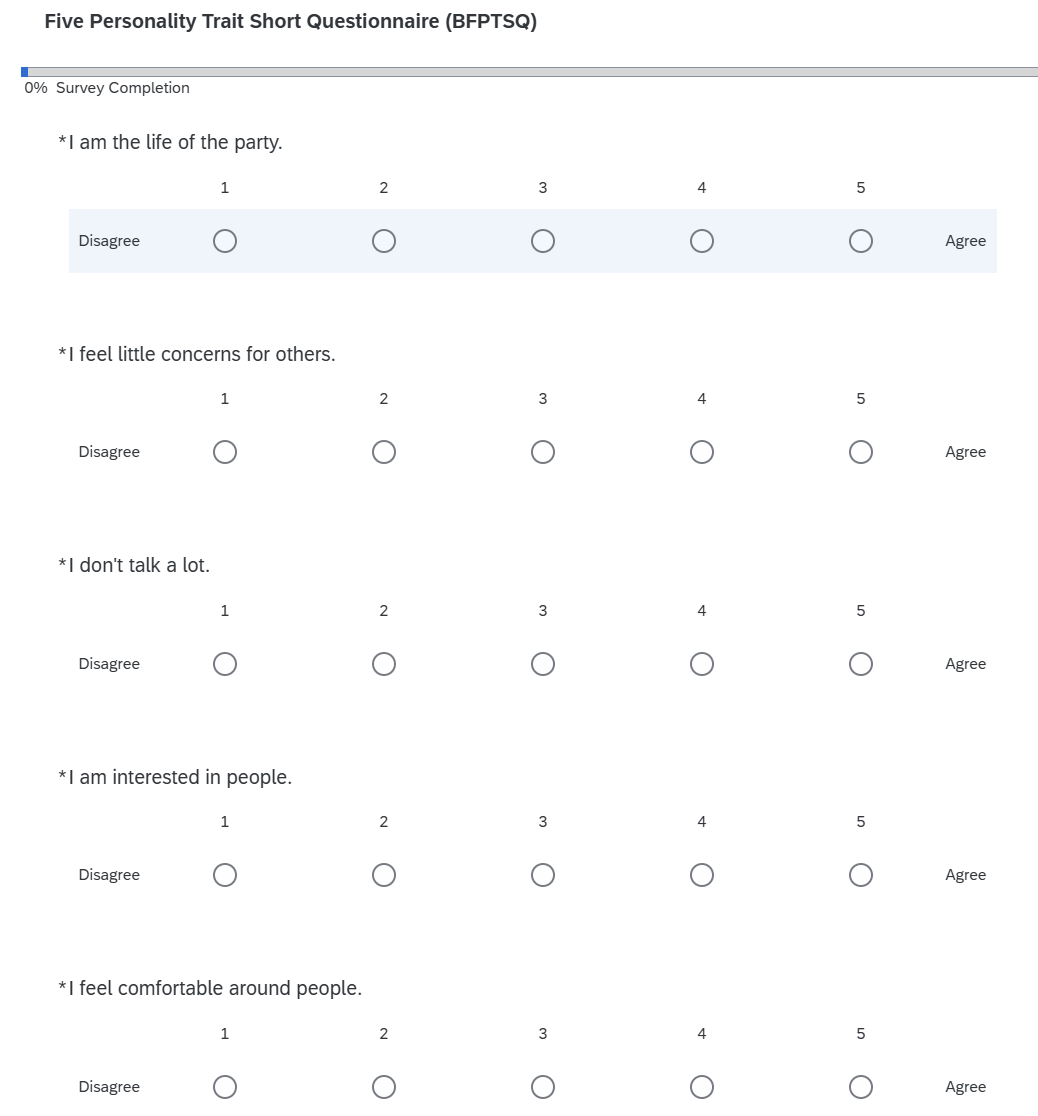}
        \caption{Page 1.}
    \end{subfigure}
    \hfill
    \begin{subfigure}[t]{0.48\textwidth}
        \centering
        \includegraphics[width=\linewidth]{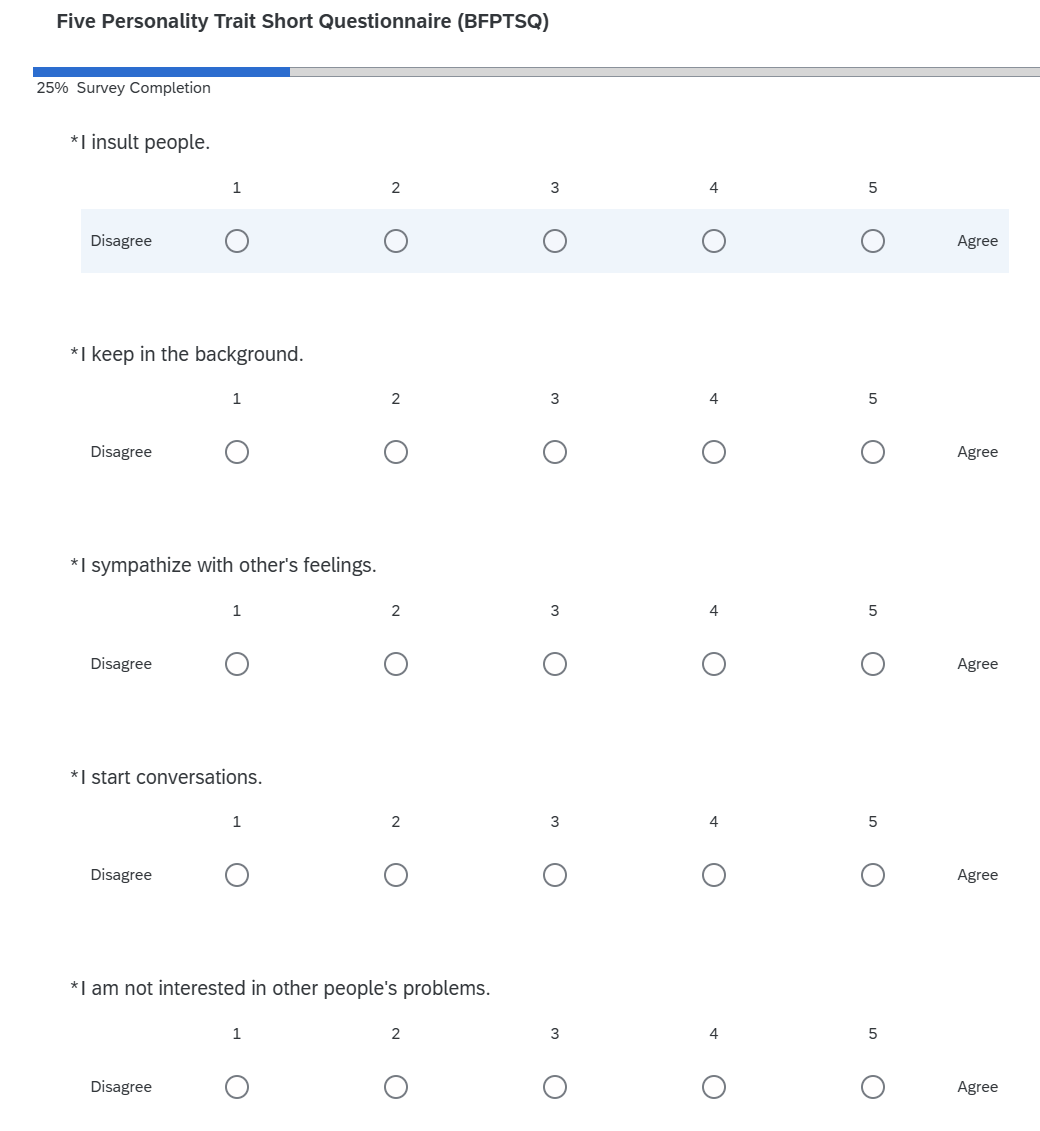}
        \caption{Page 2.}
    \end{subfigure}

    \vspace{0.8em}

    \begin{subfigure}[t]{0.48\textwidth}
        \centering
        \includegraphics[width=\linewidth]{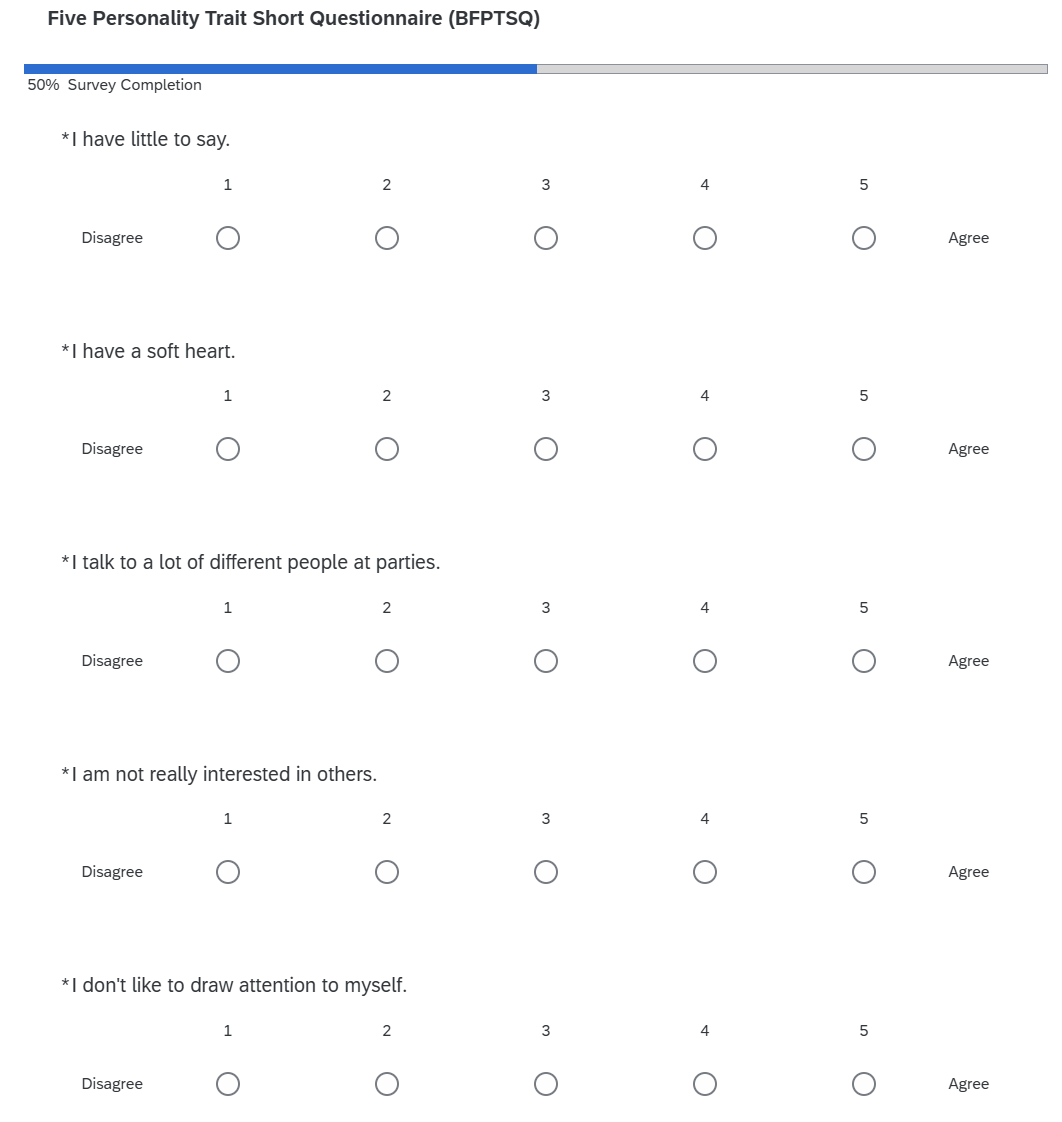}
        \caption{Page 3.}
    \end{subfigure}
    \hfill
    \begin{subfigure}[t]{0.48\textwidth}
        \centering
        \includegraphics[width=\linewidth]{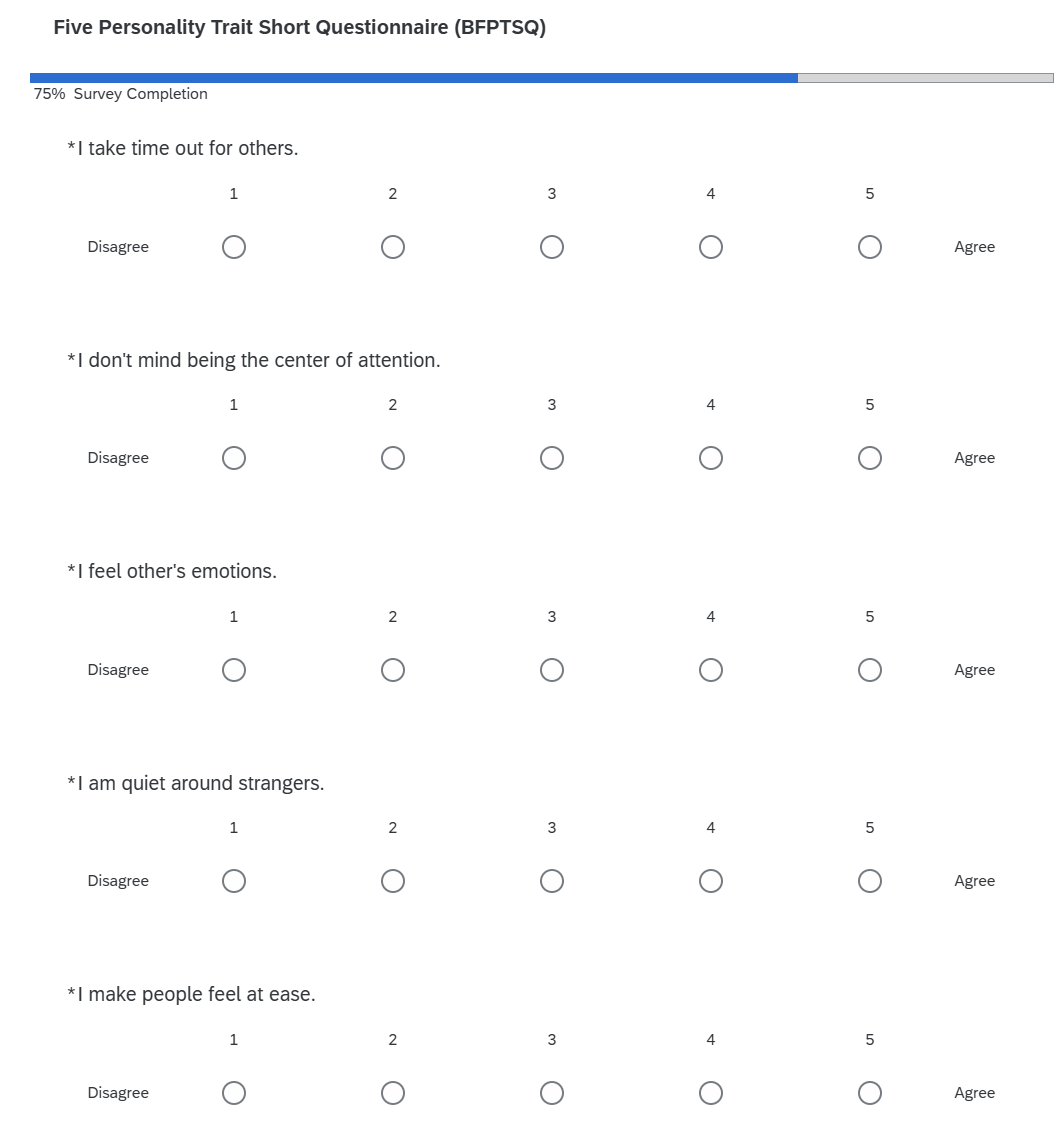}
        \caption{Page 4.}
    \end{subfigure}

    \caption{Personality questionnaire used in the human study. These responses are collected as supplementary information for analyzing whether participant personality traits align with behavior patterns observed in the synthetic personality-conditioned AI evaluation.}
    \label{fig:appendix_personality}
\end{figure*}

\begin{figure}[t]
    \centering
    \includegraphics[width=\linewidth]{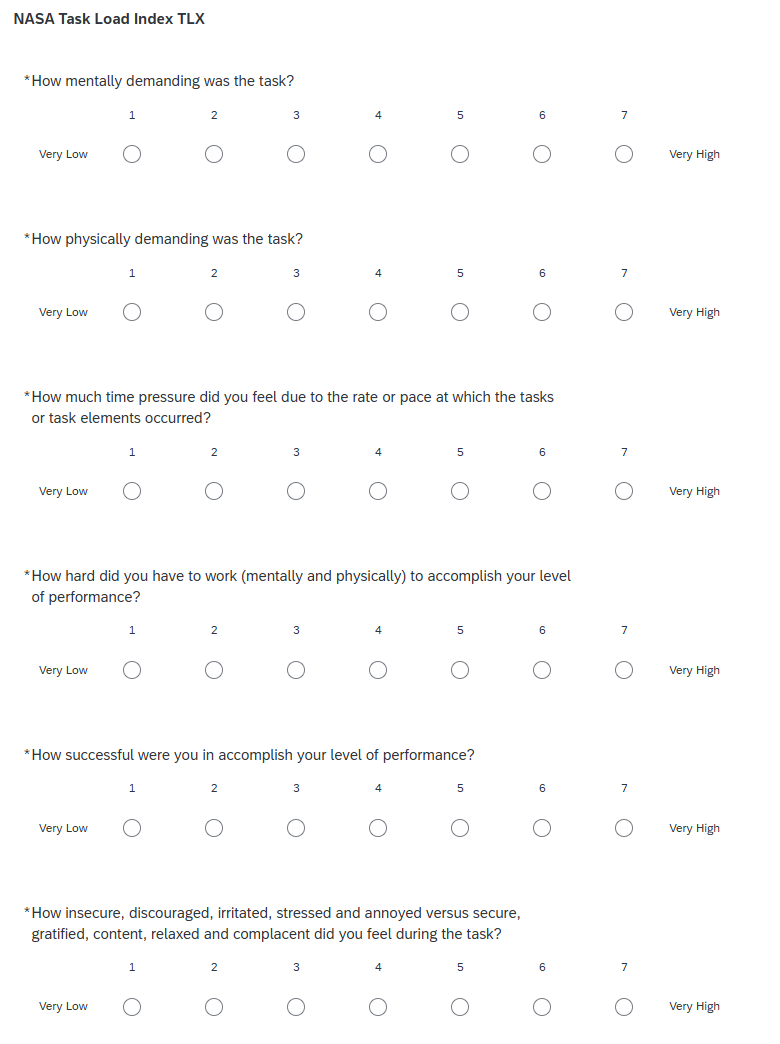}
    \caption{NASA-TLX workload questionnaire used to measure participants' perceived workload during the human study.}
    \label{fig:appendix_nasa_tlx}
\end{figure}

The questionnaire battery consisted of three parts: a team-effectiveness questionnaire, a personality questionnaire, and a workload questionnaire based on NASA-TLX. The team-effectiveness questionnaire, shown in Figure~\ref{fig:appendix_teq}, measured perceived team fluency, trust, work balance, and satisfaction using 5-point Likert-scale items. The personality questionnaire, shown in Figure~\ref{fig:appendix_personality}, was included as supplementary information for future analysis of whether real participant personality traits align with or help explain the personality-conditioned synthetic AI evaluations. The NASA-TLX questionnaire, shown in Figure~\ref{fig:appendix_nasa_tlx}, measured perceived workload during the task.

\begin{figure}[t]
    \centering
    \includegraphics[width=\linewidth]{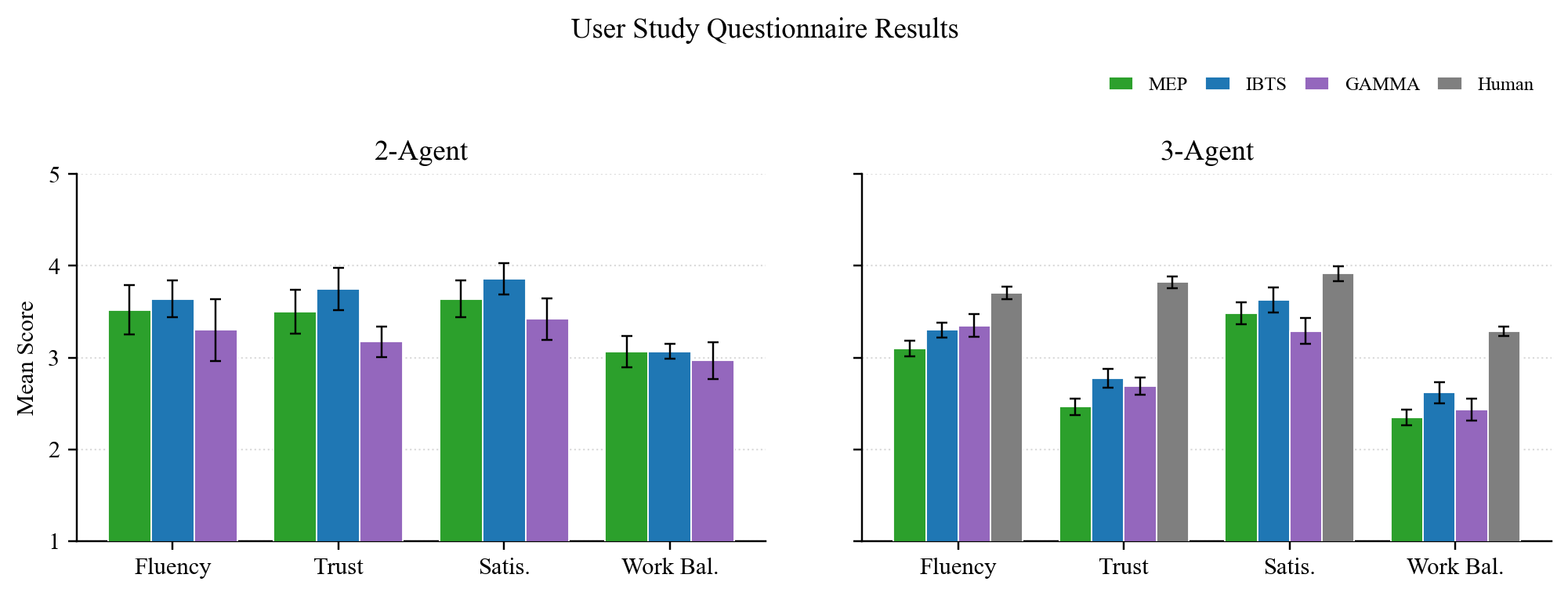}
    \caption{Post-game questionnaire ratings averaged over fluency, trust, satisfaction, and work balance. The 3-agent plot includes human teammates as a reference.}
    \label{fig:survey_appendix}
    \vspace{-3mm}
\end{figure}

\subsection{Post-Game Questionnaire Results}
\label{app:human_study_survey}

In addition to task scores, participants completed post-game questionnaires assessing perceived teammate quality along four dimensions: fluency, trust, satisfaction, and work balance. The fluency items are adapted from prior work on evaluating fluency in human--robot collaboration~\cite{hoffman2019evaluating}, while the trust and satisfaction items extend the user-study questionnaire design used in TALENTS~\cite{li2025adaptively}. Figure~\ref{fig:survey_appendix} reports the average ratings for each AI teammate across the 2-agent and 3-agent human-study settings. In the 3-agent setting, ratings for the human teammate are also included as a reference point.

The questionnaire results provide complementary evidence about subjective teammate perception. While human partners remain a strong subjective reference in group play, the learned-agent ratings suggest that IBTS is generally perceived favorably relative to the learned baselines across several dimensions. These results support the main task-score findings by showing that the performance improvements of IBTS do not come at the cost of substantially degraded perceived teammate quality.

\section{Societal Impact}
\label{app:societal_impact}

This work studies how to train machine teammates that can coordinate with diverse human partners as mixed HMT scale beyond dyadic interaction. A positive impact of this research is that it may help future assistive agents, robots, or collaborative decision-support systems support groups of people in shared tasks, especially when communication is limited or team composition changes.

At the same time, systems that steer team coordination may also shape human behavior in unintended ways. A poorly aligned AI teammate could disrupt human-human coordination, over-optimize for task reward at the expense of user preferences, or encourage interaction patterns that are efficient but not desirable for all participants. Our method also does not guarantee alignment with human values or individual preferences. Before deployment in real-world settings, such agents should be evaluated with human preference data, safety constraints, and domain-specific oversight to ensure that coordination improvements do not come at the cost of user autonomy, fairness, or trust.


\end{document}